\documentclass{article} 
\usepackage{iclr2019_conference,times}

\usepackage{amsmath,amsfonts,bm}

\def\eqref#1{equation~\ref{#1}}

\def\1{\bm{1}}

\DeclareMathAlphabet{\mathsfit}{\encodingdefault}{\sfdefault}{m}{sl}
\SetMathAlphabet{\mathsfit}{bold}{\encodingdefault}{\sfdefault}{bx}{n}

\usepackage{hyperref}
\usepackage{url}
\usepackage{graphicx}
\usepackage[title]{appendix}

\usepackage{color}
\usepackage{verbatim}
\usepackage{amsmath,amssymb,amsfonts}
\usepackage{algorithm2e}
\usepackage{multirow}

\title{Dynamic Sparse Graph for Efficient Deep Learning}

\author{Liu Liu$^{12*}$, Lei Deng$^{2*}$, Xing Hu$^2$, Maohua Zhu$^2$, Guoqi Li$^3$, Yufei Ding$^2$, Yuan Xie$^1$\\
$^1$Department of Electrical and Computer Engineering, University of California, Santa Barbara\\
$^2$Department of Computer Science, University of California, Santa Barbara\\
$^3$Center for Brain Inspired Computing Research, \\
$~~$Department of Precision Instrument, Tsinghua University\\
$^*$Equal contribution\\
\texttt{\{liu\_liu, leideng, huxing, maohua, yuanxie\}@ece.ucsb.edu}\\
\texttt{yufeiding@cs.ucsb.edu}\\
\texttt{liguoqi@mail.tsinghua.edu.cn}\\
}

\iclrfinalcopy 
\begin{document}

\maketitle

\begin{abstract}
We propose to execute deep neural networks (DNNs) with dynamic and sparse graph (DSG) structure for compressive memory and accelerative execution during both training and inference. The great success of DNNs motivates the pursuing of lightweight models for the deployment onto embedded devices. However, most of the previous studies optimize for inference while neglect training or even complicate it. Training is far more intractable, since (i) the neurons dominate the memory cost rather than the weights in inference; (ii) the dynamic activation makes previous sparse acceleration via one-off optimization on fixed weight invalid; (iii) batch normalization (BN) is critical for maintaining accuracy while its activation reorganization damages the sparsity. To address these issues, DSG activates only a small amount of neurons with high selectivity at each iteration via a dimension-reduction search and obtains the BN compatibility via a double-mask selection. Experiments show significant memory saving (1.7-4.5x) and operation reduction (2.3-4.4x) with little accuracy loss on various benchmarks.
\end{abstract}

\section{Introduction}
Deep Neural Networks (DNNs) \citep{lecun2015deep} have been achieving impressive progress in a wide spectrum of domains \citep{simonyan2014very,he2016deep,abdel2014convolutional, redmon2016yolo9000,wu2016google}, while the models are extremely memory- and compute-intensive. The high representational and computational costs motivate many researchers to investigate approaches on improving the execution performance, including matrix or tensor decomposition \citep{xue2014singular,novikov2015tensorizing, garipov2016ultimate,yang2017tensor,alvarez2017compression}, data quantization \citep{courbariaux2016binarized,zhou2016dorefa,deng2018gxnor,leng2017extremely,wen2017terngrad,wu2018training,mckinstry2018discovering}, and network pruning \citep{ardakani2016sparsely,han2015learning,han2015deep,liu2017learning,li2016pruning, he2017channel,luo2017thinet,wen2016learning,molchanov2016pruning,sun2017meprop, spring2017scalable,lin2017predictivenet,zhang2018adam,he2018soft,chin2018layer, ye2018rethinking,luo2018autopruner,hu2018novel,he2018amc}. However, most of the previous work aim at inference while the challenges for reducing the representational and computational costs of training are not well-studied. Although some works demonstrate acceleration in the distributed training \citep{lin2017deep,goyal2017accurate,you2017imagenet}, we target at the single-node optimization, and our method can also boost training in a distributed fashion.

\begin{figure}
\centering
\includegraphics[width=0.97\textwidth]{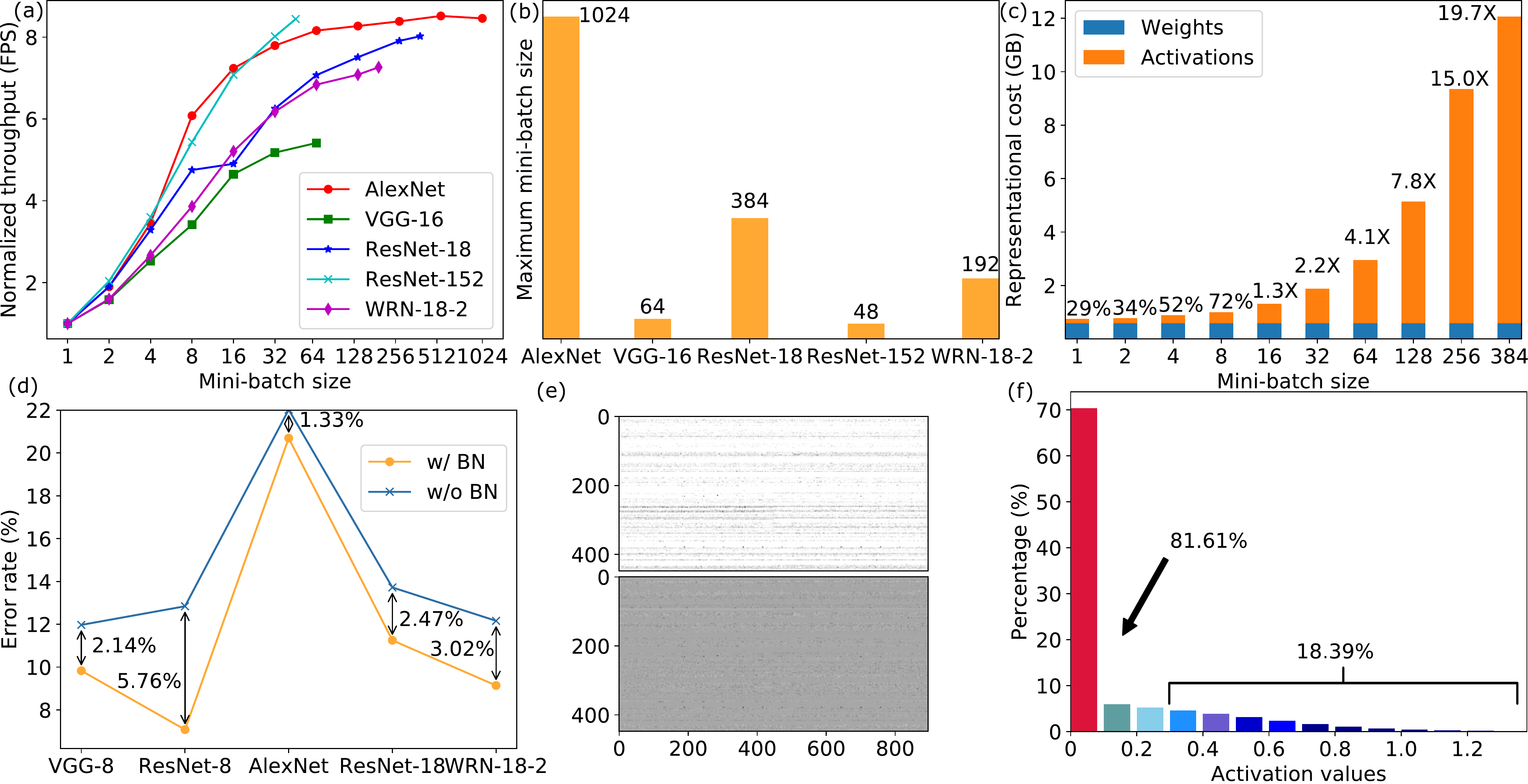}
\vspace{-5pt}
\caption{Comprehensive motivation illustration. (a) Using larger mini-batch size helps improve throughput until it is compute-bound; (b) Limited memory capacity on a single computing node prohibits the use of large mini-batch size; (c) Neuronal activation dominates the representational cost when mini-batch size becomes large; (d) BN is indispensable for maintaining accuracy; (e) Upper and lower one are the feature maps before and after BN, respectively. However, using BN damages the sparsity through information fusion; (f) There exists such great representational redundancy that more than 80\% of activations are close to zero.}
\vspace{-10pt}
\label{Motivation}
\end{figure}

DNN training, which demands much more hardware resources in terms of both memory capacity and computation volume, is far more challenging than inference. Firstly, activation data in training will be stored for backpropagation, significantly increasing the memory consumption. Secondly, training iteratively updates model parameters using mini-batched stochastic gradient descent. We almost always expect larger mini-batches for higher throughput (Figure \ref{Motivation}(a)), faster convergence, and better accuracy \citep{smith2017don}. However, memory capacity is often the limitation factor (Figure \ref{Motivation}(b)) that may cause performance degradation or even make large models with deep structures or targeting high-resolution vision tasks hard to train \citep{he2016deep,wu2018group}. 

It is difficult to apply existing sparsity techniques towards inference phase to training phase because of the following reasons: 1) Prior arts mainly compress the pre-trained and fixed weight parameters to reduce the off-chip memory access in inference \citep{han2016eie,han2017ese}, while instead, the dynamic neuronal activations turn out to be the crucial bottleneck in training \citep{jain2018gist}, making the prior inference-oriented methods inefficient. Besides, during training we need to stash a vast batched activation space for the backward gradient calculation. Therefore, neuron activations creates a new memory bottleneck (Figure \ref{Motivation}(c)). In this paper, we will sparsify the neuron activations for training compression. 2) The existing inference accelerations usually add extra optimization problems onto the critical path \citep{wen2016learning,molchanov2016pruning,liu2017learning,luo2017thinet,liang2018crossbar,zhang2018adam,hu2018novel,luo2018autopruner, ye2018rethinking}, i.e., `complicated training $\Rightarrow$ simplified inference', which embarrassingly complicates the training phase. 3) Moreover, previous studies reveal that batch normalization (BN) is crucial for improving accuracy and robustness (Figure \ref{Motivation}(d)) through activation fusion across different samples within one mini-batch for better representation \citep{morcos2018importance,ioffe2015batch}. BN almost becomes a standard training configuration; however, inference-oriented methods seldom discuss BN and treat BN parameters as scaling and shift factors in the forward pass. We further find that BN will damage the sparsity due to the activation reorganization (Figure \ref{Motivation}(e)). Since this work targets both training and inference, the BN compatibility problem should be addressed.

From the view of information representation, the activation of each neuron reflects its selectivity to the current stimulus sample \citep{morcos2018importance}, and this selectivity dataflow propagates layer by layer forming different representation levels. Fortunately, there is much representational redundancy, for example, lots of neuron activations for each stimulus sample are so small and can be removed (Figure \ref{Motivation}(f)). Motivated by above comprehensive analysis regarding memory and compute, we propose to search critical neurons for constructing a sparse graph at every iteration. By activating only a small amount of neurons with a high selectivity, we can significantly save memory and simplify computation with tolerable accuracy degradation. Because the neuron response dynamically changes under different stimulus samples, the sparse graph is variable. The neuron-aware dynamic and sparse graph (DSG) is fundamentally distinct from the static one in previous work on permanent weight pruning since we never prune the graph but activate part of them each time. Therefore, we maintain the model expressive power as much as possible. A graph selection method, dimension-reduction search, is designed for both compressible activations with element-wise unstructured sparsity and accelerative vector-matrix multiplication (VMM) with vector-wise structured sparsity. Through double-mask selection design, it is also compatible with BN. We can use the same selection pattern and extend our method to inference. In a nutshell, we propose a compressible and accelerative DSG approach supported by dimension-reduction search and double-mask selection. It can achieve 1.7-4.5x memory compression and 2.3-4.4x computation reduction with minimal accuracy loss. This work simultaneously pioneers the approach towards efficient online training and offline inference, which can benefit the deep learning in both the cloud and the edge.

\section{Approach}
\begin{figure}[!htbp]
\centering
\includegraphics[width=0.97\textwidth]{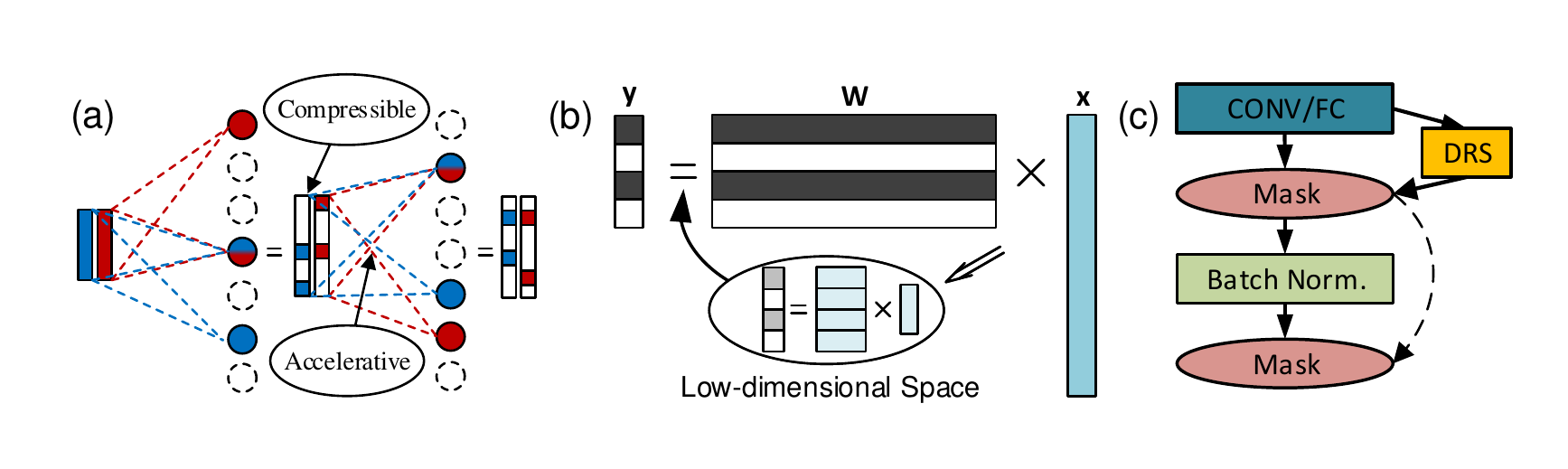}
\vspace{-10pt}
\caption{(a) Illustration of dynamic and sparse graph (DSG); (b) Dimension-reduction search for construction of DSG; (c) Double-mask selection for BN compatibility. `DRS' denotes dimension-reduction search.}
\label{Approach}
\end{figure}

Our method forms DSGs for different inputs, which are accelerative and compressive, as shown in Figure\ref{Approach}(a). On the one hand, choosing a small number of critical neurons to participate in computation, DSG can reduce the computational cost by eliminating calculations of non-critical neurons. On the other hand, it can further reduce the representational cost via compression on sparsified activations.
Different from previous methods using permanent pruning, our approach does not prune any neuron and the associated weights; instead, it activates a sparse graph according to the input sample at each iteration. Therefore, DSG does not compromise the expressive power of the model. 

\vspace{-12pt}

Constructing DSG needs to determine which neurons are critical. A naive approach is to select critical neurons according to the output activations. If the output neurons have a small or negative activation value, i.e., not selective to current input sample, they can be removed for saving representational cost. Because these activations will be small or absolute zero after the following ReLU non-linear function (i.e., ReLU($x$) $=$ max(0, $x$)), it's reasonable to set all of them to be zero. However, this naive approach requires computations of all VMM operations within each layer before the selection of critical neurons, which is very costly.

\subsection{Dimension-reduction Search}
To avoid the costly VMM operations in the mentioned naive selection, we propose an efficient method, i.e., dimension reduction search, to estimate the importance of output neurons. As shown in Figure\ref{Approach}(b), we first reduce the dimensions of $\textbf{X}$ and $\textbf{W}$, and then execute the lightweight VMM operations in a low-dimensional space with minimal cost. After that, we estimate the neuron importance according to the virtual output activations. Then, a binary selection mask can be produced in which the zeros represent the non-critical neurons with small activations that are removable. We use a top-\textit{k} search method that only keeps largest \textit{k} neurons, where an inter-sample threshold sharing mechanism is leveraged to greatly reduce the search cost \footnote{Implementation details are shown in Appendix B.}. Note that \textit{k} is determined by the output size and a pre-configured sparsity parameter $\gamma$.  Then we can just compute the accurate activations of the critical neurons in the original high-dimensional space and avoid the calculation of the non-critical neurons. Thus, besides the compressive sparse activations, the dimension-reduction search can further save a significant amount of expensive operations in the high-dimensional space.

\begin{figure}[!htbp]
\centering
\includegraphics[width=0.97\textwidth]{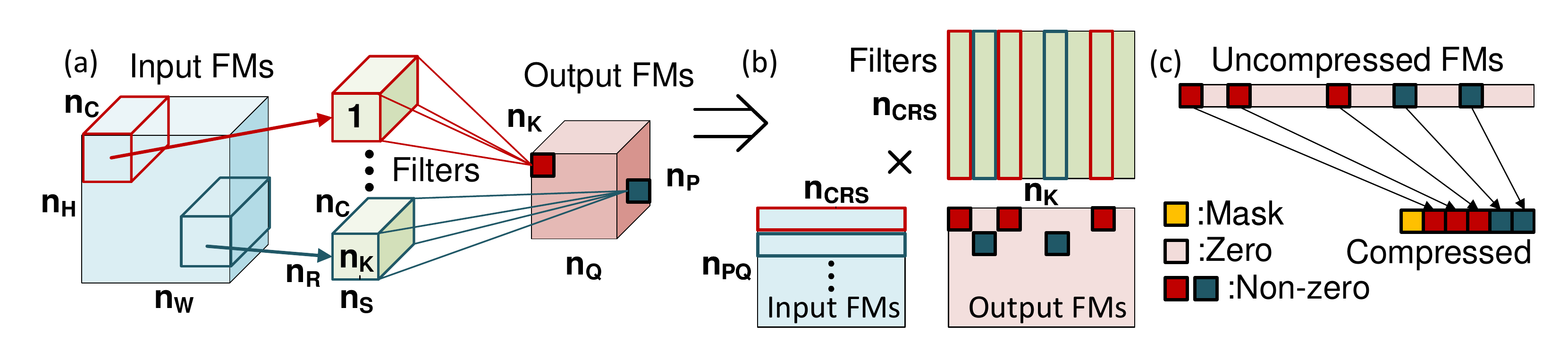}
\vspace{-12pt}
\caption{Compressive and accelerative DSG. (a) Original dense convolution; (b) Converted accelerative VMM operation; (c) Zero-value compression.}
\label{Conv}
\end{figure}

In this way, a vector-wise structured sparsity can be achieved, as shown in Figure \ref{Conv}(b). The ones in the selection mask (marked as colored blocks) denote the critical neurons, and the non-critical ones can bypass the memory access and computation of their corresponding columns in the weight matrix. Furthermore, the generated sparse activations can be compressed via the zero-value compression \citep{zhang2000frequent,vijaykumar2015case,rhu2018compressing} (Figure \ref{Conv}(c)). Consequently, it is critical to reduce the vector dimension but keep the activations calculated in the low-dimensional space as accurate as possible, compared to the ones in the original high-dimensional space.

\subsection{Sparse Random Projection for Efficient Dimension-reduction Search}
\underline{\emph{Notations}:} Each CONV layer has a four dimensional weight tensor ($n_K$, $n_C$, $n_R$, $n_S$), where $n_K$ is the number of filters, i.e., the number of output feature maps (FMs); $n_C$ is the number of input FMs; ($n_R$, $n_S$) represents the kernel size. Thus, the CONV layer in Figure \ref{Conv}(a) can be converted to many VMM operations, as shown in Figure \ref{Conv}(b). Each row in the matrix of input FMs is the activations from a sliding window across all input FMs ($n_{CRS}=n_C\times n_R\times n_S$), and after the VMM operation with the weight matrix ($n_{CRS}\times n_K$) it can generate $n_K$ points at the same location across all output FMs. Further considering the $n_{PQ}=n_P\times n_Q$ size of each output FM and the mini-batch size of $m$, the whole $n_{PQ}\times m$ rows of VMM operations has a computational complexity of $O(m\times n_{PQ}\times n_{CRS}\times n_K)$. For the FC layer with $n_C$ input neurons and $n_K$ output neurons, this complexity is $O(m\times n_C\times n_K)$. Note that here we switch the order of BN and ReLU layer from `CONV/FC-BN-ReLU' to `CONV/FC-ReLU-BN', because it's hard to determine the activation value of the non-critical neurons if the following layer is BN (this value is zero for ReLU). As shown in previous work, this reorganization could bring better accuracy \citep{mishkin2015all}. 

For the sake of simplicity, we just consider the operation for each sliding window in the CONV layer or the whole FC layer under one single input sample as a basic optimization problem. The generation of each output activation $y_j$ requires an inner product operation, as follows:
\begin{equation}
y_j=\varphi(\langle \textbf{X}_i, \textbf{W}_j \rangle)
\label{Inner}
\end{equation}
where $\textbf{X}_i$ is the $i$-th row in the matrix of input FMs (for the FC layer, there is only one $\textbf{X}$ vector), $\textbf{W}_j$ is the $j$-th column of the weight matrix $W$, and $\varphi(\cdot)$ is the neuronal transformation (e.g., ReLU function, here we abandon bias). Now, according to equation (\ref{Inner}), the preservation of the activation is equivalent to preserve the inner product.

We introduce a dimension-reduction lemma, named Johnson-Lindenstrauss Lemma (JLL) \citep{johnson1984extensions}, to implement the dimension-reduction search with inner product preservation. This lemma states that a set of points in a high-dimensional space can be embedded into a low-dimensional space in such a way that the Euclidean distances between these points are nearly preserved. Specifically, given $0<\epsilon<1$, a set of $N$ points in $\mathbb{R}^d$ (i.e., all $\textbf{X}_i$ and $\textbf{W}_j$), and a number of $k>O(\frac{log(N)}{\epsilon ^2})$, there exists a linear map $f: \mathbb{R}^d\Rightarrow \mathbb{R}^k$ such that
\begin{equation}
(1-\epsilon)\Vert \textbf{X}_i - \textbf{W}_j \Vert ^2 \leq \Vert f(\textbf{X}_i) - f(\textbf{W}_j) \Vert ^2 \leq (1+\epsilon)\Vert \textbf{X}_i - \textbf{W}_j \Vert ^2
\label{JL_distance}
\end{equation}
for any given $\textbf{X}_i$ and $\textbf{W}_j$ pair, where $\epsilon$ is a hyper-parameter to control the approximation error, i.e., larger $\epsilon$ $\Rightarrow$ larger error. When $\epsilon$ is sufficiently small, one corollary from JLL is the following norm preservation \citep{vu2016random, Kakade_cmsc35900}:
\begin{equation}
P[~(1-\epsilon)\Vert \textbf{Z}\Vert ^2 \leq \Vert f(\textbf{Z})\Vert ^2 \leq (1+\epsilon)\Vert \textbf{Z}\Vert ^2~]\geq 1-O(\epsilon ^2)
\label{eq:JL_norm}
\end{equation}
where $\textbf{Z}$ could be any $\textbf{X}_i$ or $\textbf{W}_j$, and $P$ denotes a probability. It means the vector norm can be preserved with a high probability controlled by $\epsilon$. Given these basics, we can further get the inner product preservation:
\begin{equation}
P[~| \langle f(\textbf{X}_i), f(\textbf{W}_j) \rangle - \langle \textbf{X}_i, \textbf{W}_j \rangle |\leq \epsilon~]\geq 1-O(\epsilon ^2).
\label{eq:JL_inner}
\end{equation}
The detailed proof can be found in Appendix \ref{app:proof}. 

Random projection \citep{vu2016random, ailon2009fast, achlioptas2001database} is widely used to construct the linear map $f(\cdot)$. Specifically, the original $d$-dimensional vector is projected to a $k$-dimensional ($k \ll d$) one, using a random $k\times d$ matrix $\textbf{R}$. Then we can reduce the dimension of all $\textbf{X}_i$ and $\textbf{W}_j$ by
\begin{equation}
f(\textbf{X}_i) = \frac{1}{\sqrt{k}}\textbf{R}\textbf{X}_i\in \mathbb{R}^k,~~f(\textbf{W}_j) = \frac{1}{\sqrt{k}}\textbf{R}\textbf{W}_j\in \mathbb{R}^k.
\label{eq:DRS}
\end{equation}
The random projection matrix $\textbf{R}$ can be generated from Gaussian distribution \citep{ailon2009fast}. In this paper, we adopt a simplified version, termed as sparse random projection \citep{achlioptas2001database, bingham2001random, li2006very} with
\begin{equation}
P(\textbf{R}_{pq}=\sqrt{s})=\frac{1}{2s};~~P(\textbf{R}_{pq}=0)=1-\frac{1}{s};~~P(\textbf{R}_{pq}=-\sqrt{s})=\frac{1}{2s}
\label{eq:Projection}
\end{equation}
for all elements in $\textbf{R}$. This $\textbf{R}$ only has ternary values that can remove the multiplications during projection, and the remained additions are very sparse. Therefore, the projection overhead is negligible compared to other high-precision operations involving multiplication. Here we set $s=3$ with 67\% sparsity in statistics.

\begin{figure}[!htbp]
\centering
\includegraphics[width=0.9\textwidth]{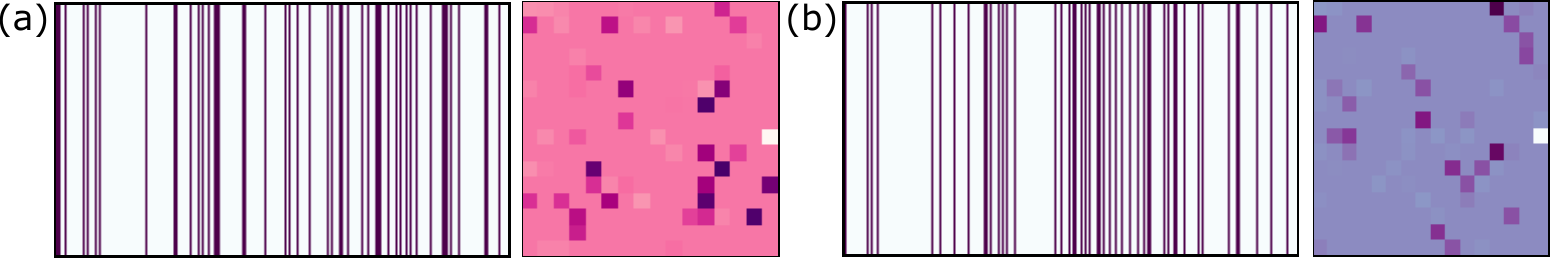}
\caption{Structured selection via dynamic dimension-reduction search for producing sparse pattern of neuronal activations.}
\label{Dynamic}
\end{figure}

Equation (\ref{eq:JL_inner}) indicates the low-dimensional inner product $\langle f(\textbf{X}_i), f(\textbf{W}_j) \rangle$ can still approximate the original high-dimensional one $\langle \textbf{X}_i, \textbf{W}_j \rangle$ in equation (\ref{Inner}) if the reduced dimension is sufficiently high. Therefore, it is possible to calculate equation (\ref{Inner}) in a low-dimensional space for activation estimation, and select the important neurons. As shown in Figure \ref{Conv}(b), each sliding window dynamically selects its own important neurons for the calculation in high-dimensional space, marked in red and blue as two examples. Figure \ref{Dynamic} visualizes two sliding windows in a real network to help understand the dynamic process of dimension-reduction search. Here the neuronal activation vector ($n_K$ length) is reshaped to a matrix for clarity. Now For the CONV layer, the computational complexity is only $O[~m\times n_{PQ}\times n_K\times(k+ (1-\gamma) \times n_{CRS})~]$, which is less than the original high-dimensional computation with $O(m\times n_{PQ}\times n_{CRS}\times n_K)$ complexity because we usually have $[~k+(1-\gamma) \times n_{CRS}~]\ll n_{CRS}$. For the FC layer, we also have $O[~m\times n_K\times(k+ (1-\gamma) \times n_C)~]\ll O(m\times n_C\times n_K)$.

\subsection{Double-mask Selection for BN Compatibility}
To deal with the important but intractable BN layer, we propose a double-mask selection method presented in Figure \ref{Approach}(c). After the dimension-reduction search based importance estimation, we produce a sparsifying mask that removes the unimportant neurons. The ReLU activation function can maintain this mask by inhibiting the negative activation (actually all the activations of the CONV layer or FC layer after the selection mask are positive with reasonably large sparsity). However, the BN layer will damage this sparsity through inter-sample activation fusion. To address this issue, we copy the same selection mask before the BN layer and directly use it on the BN output. It is straightforward but reasonable because we find that although BN causes the zero activation to be non-zero (Figure \ref{Motivation}(f)), these non-zero activations are still very small and can also be removed. This is because BN just scales and shifts the activations that won't change the relative sort order. In this way, we can achieve fully sparse activation dataflow. The back propagated gradients will also be forcibly sparsified every time they pass a mask layer.
\section{Experimental Results}
\subsection{Experiment Setup}

The overall training algorithm is presented in Appendices \ref{app:DRS}. Going through the dataflow where the red color denotes the sparse tensors, a widespread sparsity in both the forward and backward passes is demonstrated. The projection matrices are fixed after a random initialization at the beginning of training. We just update the projected weights in the low-dimensional space every 50 iterations to reduce the projection overhead. The detailed search method and the computational complexity of the dimension-reduction search are provided in Appendix \ref{app:DRS}. Regarding the evaluation network models, we use LeNet \citep{lecun1998gradient} and a multi-layered perceptron (MLP) on small-scale FASHION dataset \citep{xiao2017fashion}, VGG8 \citep{courbariaux2016binarized,deng2018gxnor}/ResNet8 (a customized ResNet-variant with 3 residual blocks and 2 FC layers)/ResNet20/WRN-8-2 \citep{zagoruyko2016wide} on medium-scale CIFAR10 dataset \citep{krizhevsky2009learning}, VGG8/WRN-8-2 on another medium-scale CIFAR100 dataset \citep{krizhevsky2009learning}, and AlexNet \citep{krizhevsky2012imagenet}/VGG16 \citep{simonyan2014very}/ResNet18, ResNet152 \citep{he2016deep}/WRN-18-2 \citep{zagoruyko2016wide} on large-scale ImageNet dataset \citep{deng2009imagenet} as workloads. The programming framework is PyTorch and the training platform is based on NVIDIA Titan Xp GPU. We adopt the zero-value compression method \citep{zhang2000frequent,vijaykumar2015case,rhu2018compressing} for memory compression and MKL compute library \citep{wang2014intel}  on Intel Xeon CPU for acceleration evaluation. 

\subsection{Accuracy Analysis}
In this section, we provide a comprehensive analysis regarding the influence of sparsity on accuracy and explore the robustness of MLP and CNN, the graph selection strategy, the BN compatibility, and the importance of width and depth.

\begin{figure}[!htbp]
\centering
\includegraphics[width=0.97\textwidth]{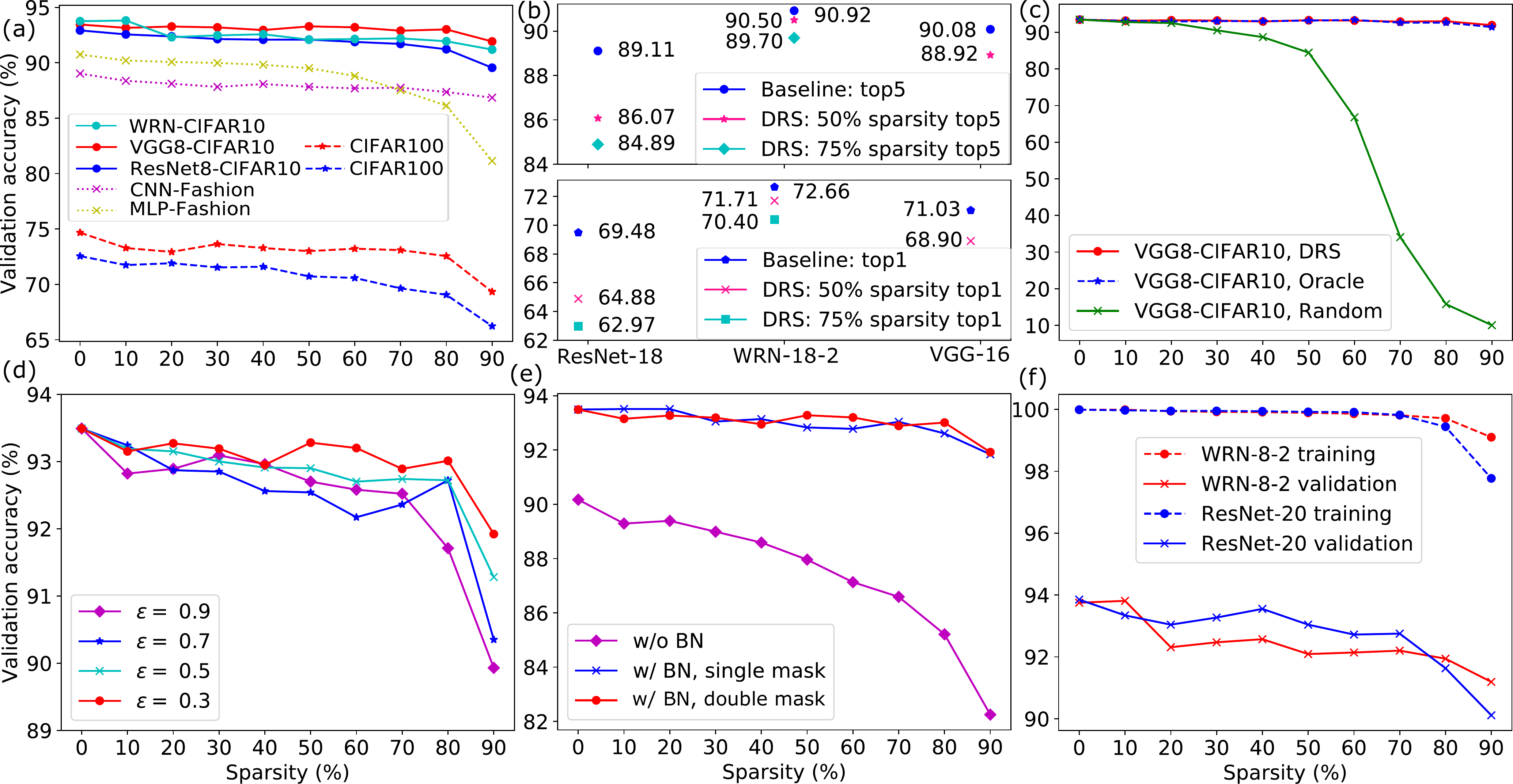}
\caption{Comprehensive analysis on sparsity v.s. accuracy. (a) \& (b) Accuracy using DSG; (c) Influence of the graph selection strategy; (d) Influence of the dimension-reduction degree; (e) Influence of the double-mask selection for BN compatibility; (f) Influence of the network depth and width. `DRS' denotes dimension-reduction search.}
\label{Accuracy}
\end{figure}

\textbf{Accuracy using DSG.} Figure \ref{Accuracy}(a) presents the accuracy curves on small and medium scale models by using DSG under different sparsity levels. Three conclusions are observed: 1) The proposed DSG affects little on the accuracy when the sparsity is $<$60\%, and the accuracy will present an abrupt descent with sparsity larger than 80\%. 2) Usually, the ResNet model family is more sensitive to the sparsity increasing due to fewer parameters than the VGG family. For the VGG8 on CIFAR10, the accuracy loss is still within 0.5\% when sparsity reaches 80\%. 3) Compared to MLP, CNN can tolerate more sparsity. Figure \ref{Accuracy}(b) further shows the results on large scale models on ImageNet. Because training large model is time costly, we only present several experimental points. Consistently, the VGG16 shows better robustness compared to the ResNet18, and the WRN with wider channels on each layer performs much better than the other two models.  We will discuss the topic of width and depth later.

\textbf{Graph Selection Strategy.} To investigate the influence of graph selection strategy, we repeat the sparsity vs. accuracy experiments on CIFAR10 under different selection methods. Two baselines are used here: the oracle one that keeps the neurons with top-k activations after the whole VMM computation at each layer, and the random one that randomly selects neurons to keep. The results are shown in Figure \ref{Accuracy}(c), in which we can see that our dimension-reduction search and the oracle one perform much better than the random selection under high sparsity condition. Moreover, dimension-reduction search achieves nearly the same accuracy with the oracle top-k selection, which indicates the proposed random projection method can find an accurate activation estimation in the low-dimensional space. In detail, Figure \ref{Accuracy}(d) shows the influence of parameter $\epsilon$ that reflects the degree of dimension reduction. Lower $\epsilon$ can approach the original inner product more accurately, that brings higher accuracy but at the cost of more computation for graph selection since less dimension reduction. With $\epsilon=0.5$, the accuracy loss is within 1\% even if the sparsity reaches 80\%.

\textbf{BN Compatibility.} Figure \ref{Accuracy}(e) focuses the BN compatibility issue. Here we use dimension-reduction search for the graph sparsifying, and compare three cases: 1) removing the BN operation and using single mask; 2) keeping BN and using only single mask (the first one in Figure \ref{Approach}(c)); 3) keeping BN and using double masks (i.e. double-mask selection). The one without BN is very sensitive to the graph ablation, which indicates the importance of BN for training. Comparing the two with BN, the double-mask selection even achieves better accuracy since the regularization effect. This observation indicates the effectiveness of the proposed double-mask selection for simultaneously recovering the sparsity damaged by the BN layer and maintaining the accuracy.

\textbf{Width or Depth.} Furthermore, we investigate an interesting comparison regarding the network width and depth, as shown in Figure \ref{Accuracy}(f). On the training set, WRN with fewer but wider layers demonstrates more robustness than the deeper one with more but slimmer layers. On the validation set, the results are a little more complicated. Under small and medium sparsity, the deeper ResNet performs better (1\%) than the wider one. While when the sparsity increases substantial ($>$75\%), WRN can maintain the accuracy better. This indicates that, in medium-sparse space, the deeper network has stronger representation ability because of the deep structure; however, in ultra-high-sparse space, the deeper structure is more likely to collapse since the accumulation of the pruning error layer by layer. In reality, we can determine which type of model to use according to the sparsity requirement. In Figure \ref{Accuracy}(b) on ImageNet, the reason why WRN-18-2 performs much better is that it has wider layers without reducing the depth.

\textbf{Convergence.} DSG does not slow down the convergence speed, which can be seen from Figure \ref{Acc_Convergence}(a)-(b) in Appendix \ref{app:convergence}. This owes to the high fidelity of inner product when we use random projection to reduce the data dimension, as shown in Figure \ref{Acc_Convergence}(c). Interestingly, Figure \ref{Mask_Convergence} (also in Appendix \ref{app:convergence}) reveals that the selection mask for each sample also converges as training goes on, however, the selection pattern varies across samples. To save the selection patterns of all samples is memory consuming, which is the reason why we do not directly suspend the selection patterns after training but still do on-the-fly dimension-reduction search in inference.

\subsection{Representational Cost Reduction}
\begin{figure}[!htbp]
\centering
\includegraphics[width=0.97\textwidth]{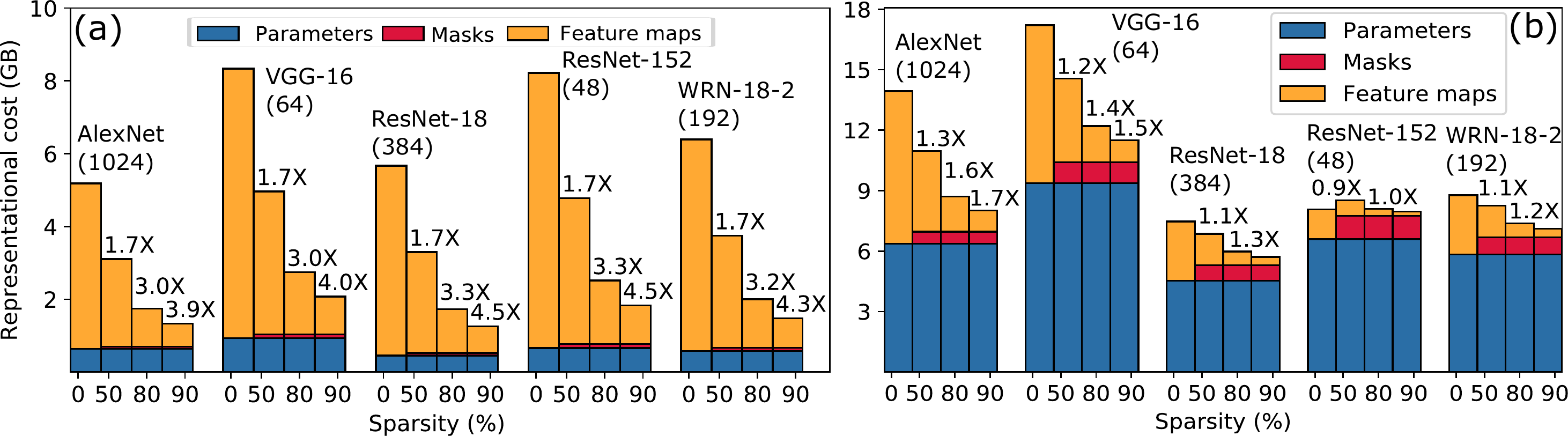}
\caption{Memory footprint comparisons for (a) training and (b) inference.}
\label{Memory1}
\end{figure}

This section presents the benefits from DSG on representational cost. We measure the memory consumption over five CNN benchmarks on both the training and inference phases. For data compression, we use zero-value compression algorithm \citep{zhang2000frequent,vijaykumar2015case,rhu2018compressing}. Figure \ref{Memory1} shows the memory optimization results, where the model name, mini-batch size, and the sparsity are provided. In training, besides the parameters, the activations across all layers should be stashed for the backward computation. Consistent with the observation mentioned above that the neuron activation beats weight to dominate memory overhead, which is different from the previous work on inference. We can reduce the overall representational cost by average 1.7x (2.72 GB), 3.2x (4.51 GB), and 4.2x (5.04 GB) under 50\%, 80\% and 90\% sparsity, respectively. If only considering the neuronal activation, these ratios could be higher up to 7.1x. The memory overhead for the selection masks is minimal ($<$2\%).

During inference, only memory space to store the parameters and the activations of the layer with maximum neuron amount is required. The benefits in inference are relatively smaller than that in training since weight is the dominant memory. On ResNet152, the extra mask overhead even offsets the compression benefit under 50\% sparsity, whereas, we can still achieve up to 7.1x memory reduction for activations and 1.7x for overall memory. Although the compression is limited for inference, it still can achieve noticeable acceleration that will be shown in the next section. Moreover, reducing costs for both training and inference is our major contribution. 

\subsection{Computational Cost Reduction}
\begin{figure}[!htbp]
\centering
\includegraphics[width=0.97\textwidth]{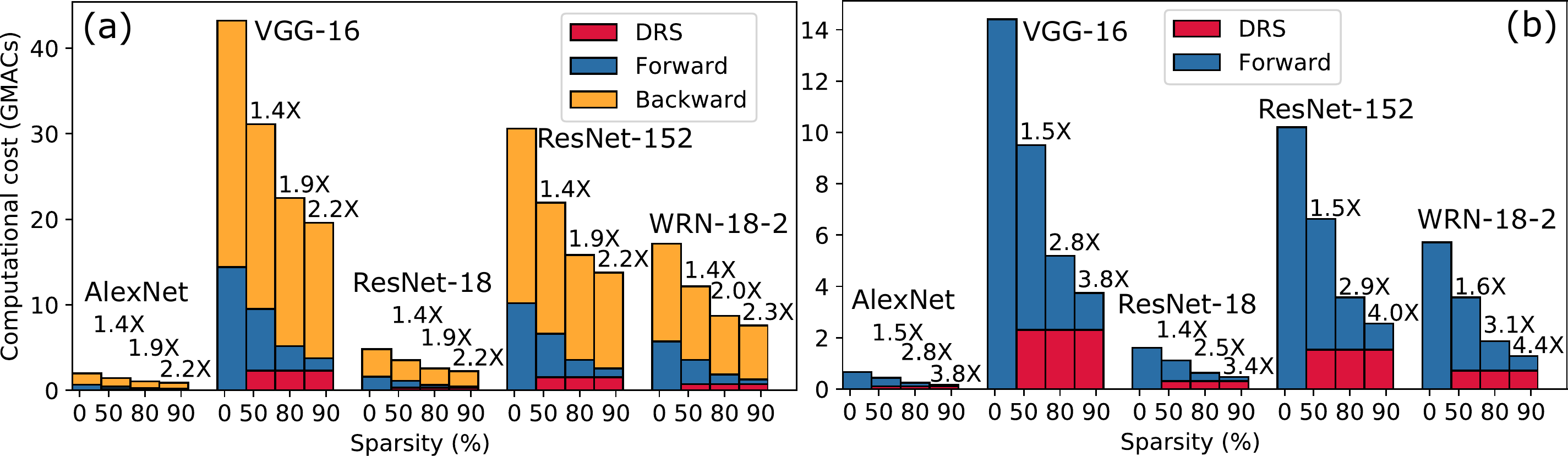}
\caption{Computational complexity comparisons for (a) training and (b) inference. `DRS' denotes dimension-reduction search.}
\label{Compute}
\end{figure}

We assess the results on reducing the computational cost for both training and inference. 
As shown in Figure \ref{Compute}, both the forward and backward pass consume much fewer operations, i.e., multiply-and-accumulate (MAC). On average, 1.4x (5.52 GMACs), 1.7x (9.43 GMACs), and 2.2x (10.74 GMACs) operation reduction are achieved in training under 50\%, 80\% and 90\% sparsity, respectively. For inference with only forward pass, the results increase to 1.5x (2.26 GMACs), 2.8x (4.22 GMACs), and 3.9x (4.87 GMACs), respectively. The overhead of the dimension-reduction search in the low-dimensional space is relatively larger  ($<$6.5\% in training and $<$19.5\% in inference) compared to the mask overhead in memory cost. Note that the training demonstrates less improvement than the inference, which is because the acceleration of the backward pass is partial. The error propagation is accelerative, but the weight gradient generation is not because of the irregular sparsity that is hard to obtain practical acceleration. Although the computation of this part is also very sparse with much fewer operations \footnote{See Algorithm \ref{algorithm} in Appendices \ref{app:DRS}}, we do not include its GMACs reduction for practical concern.

\begin{figure}[htbp]
\centering
\includegraphics[width=0.97\textwidth]{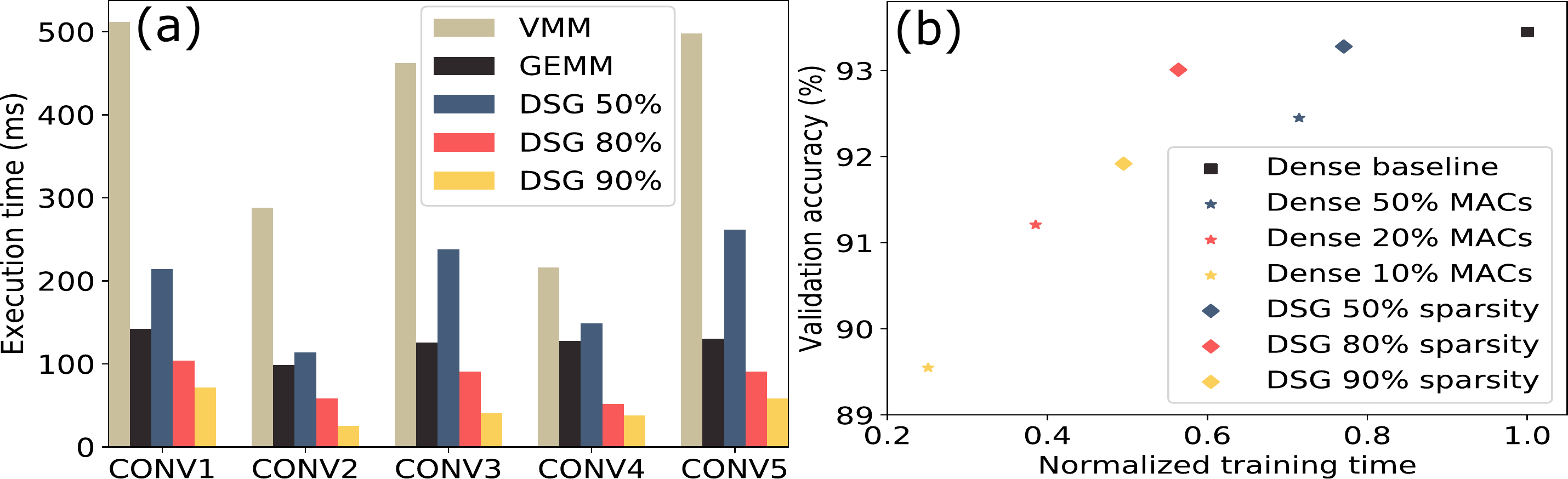}
\caption{On VGG8: (a) Layer-wise execution time comparison; (b) Validation accuracy v.s. training time of different models: large-sparse ones and smaller-dense ones with equivalent MACs.}
\label{measured}
\end{figure}

Finally, we evaluate the execution time on CPU using Intel MKL kernels (\cite{wang2014intel}).
As shown in Figure \ref{measured}(a), we evaluate the execution time of these layers after the dimension-reduction search on VGG8. Comparing to VMM baselines, our approach can achieve 2.0x, 5.0x, and 8.5x average speedup under 50\%, 80\%, and 90\% sparsity, respectively. When the baselines change to GEMM (general matrix multiplication), the average speedup decreases to 0.6x, 1.6x, and 2.7x, respectively. The reason is that DSG generates dynamic vector-wise sparsity, which is not well supported by GEMM. A potential way to improve GEMM-based implementation, at workload mapping and tiling time, is reordering executions at the granularity of vector inner-product and grouping non-redundant executions to the same tile to improve local data reuse.

On the same network, we further compare our approach with smaller dense models which could be another way to reduce the computational cost. As shown in Figure \ref{measured}(b), comparing with dense baseline, our approach can reduce training time with little accuracy loss. Even though the equivalent smaller dense models with the same effective nodes, i.e., reduced MACs, save more training time, the accuracy is much worse than our DSG approach. Figure \ref{vs_small} in Appendix \ref{app:comparison} gives more results on ResNet8 and AlexNet.

\section{Related Work}
\textbf{DNN Compression} \citep{ardakani2016sparsely} achieved up to 90\% weight sparsity by randomly removing connections. \citep{han2015learning,han2015deep} reduced the weight parameters by pruning the unimportant connections. The compression is mainly achieved on FC layers, which makes it ineffective for CONV layer-dominant networks, e.g., ResNet. To improve the pruning performance, Y. He et al. \citep{he2018amc} leveraged reinforcement learning to optimize the sparsity configuration across layers. However, it is difficult to obtain practical speedup due to the irregularity of the element-wise sparsity \citep{han2015learning,han2015deep}. Even if designing ASIC from scratch \citep{han2016eie,han2017ese}, the index overhead is enormous and it only works under high sparsity. These methods usually require a pre-trained model, iterative pruning, and fine-tune retraining, that targets inference optimization.

\textbf{DNN Acceleration} Different from compression, the acceleration work consider more on the sparse pattern. In contrast to the fine-grain compression, coarse-grain sparsity was further proposed to optimize the execution speed. Channel-level sparsity was gained by removing unimportant weight filters \citep{he2018soft,chin2018layer}, training penalty coefficients \citep{liu2017learning,ye2018rethinking,luo2018autopruner}, or solving optimization problem \citep{luo2017thinet,he2017channel,liang2018crossbar,hu2018novel}. \cite{wen2016learning} introduced a L2-norm group-lasso optimization for both medium-grain sparsity (row/column) and coarse-grain weight sparsity (channel/filter/layer). \cite{molchanov2016pruning} introduced the Taylor expansion for neuron pruning. However, they just benefit the inference acceleration, and the extra solving of the optimization problem usually makes the training more complicated. \cite{lin2017predictivenet} demonstrated predicting important neurons then bypassed the unimportant ones via low-precision pre-computation with less cost. \cite{spring2017scalable} leveraged the randomized hashing to predict the important neurons. However, the hashing search aims at finding neurons whose weight bases are similar to the input vector, which cannot estimate the inner product accurately thus will probably cause significant accuracy loss on large models. \cite{sun2017meprop} used a straightforward top-k pruning on the back propagated errors for training acceleration. But they only simplified the backward pass and presented the results on tiny FC models. Furthermore, the BN compatibility problem that is very important for large-model training still remains untouched. \cite{lin2017deep} pruned the gradients for accelerating distributed training, but the focus is on multi-node communication rather than the single-node scenario discussed in this paper. 

\section{Conclusion}
In this work, we propose DSG (dynamic and sparse graph) structure for efficient DNN training and inference through a dimension-reduction search based sparsity forecast for compressive memory and accelerative execution and a double-mask selection for BN compatibility without sacrificing model's expressive power. It can be easily extended to the inference by using the same selection pattern after training. Our experiments over various benchmarks demonstrate significant memory saving (up to 4.5x for training and 1.7x for inference) and computation reduction (up to 2.3x for training and 4.4x for inference). Through significantly boosting both forward and backward passes in training, as well as in inference, DSG promises efficient deep learning in both the cloud and edge.

\section*{Acknowledgment}
This work was partially supported by the National Science Foundations(NSF) under Grant No. 1725447 and 1730309, the National Natural Science Foundation of China under Grant No. 61603209 and 61876215. Financial support from the Beijing Innovation Center for Future Chip is also gratefully acknowledged.

\bibliography{ref}

\begin{thebibliography}{66}
\providecommand{\natexlab}[1]{#1}
\providecommand{\url}[1]{\texttt{#1}}
\expandafter\ifx\csname urlstyle\endcsname\relax
  \providecommand{\doi}[1]{doi: #1}\else
  \providecommand{\doi}{doi: \begingroup \urlstyle{rm}\Url}\fi

\bibitem[Abdel-Hamid et~al.(2014)Abdel-Hamid, Mohamed, Jiang, Deng, Penn, and
  Yu]{abdel2014convolutional}
Ossama Abdel-Hamid, Abdel-rahman Mohamed, Hui Jiang, Li~Deng, Gerald Penn, and
  Dong Yu.
\newblock Convolutional neural networks for speech recognition.
\newblock \emph{IEEE/ACM Transactions on audio, speech, and language
  processing}, 22\penalty0 (10):\penalty0 1533--1545, 2014.

\bibitem[Achlioptas(2001)]{achlioptas2001database}
Dimitris Achlioptas.
\newblock Database-friendly random projections.
\newblock In \emph{Proceedings of the twentieth ACM SIGMOD-SIGACT-SIGART
  symposium on Principles of database systems}, pp.\  274--281. ACM, 2001.

\bibitem[Ailon \& Chazelle(2009)Ailon and Chazelle]{ailon2009fast}
Nir Ailon and Bernard Chazelle.
\newblock The fast johnson--lindenstrauss transform and approximate nearest
  neighbors.
\newblock \emph{SIAM Journal on computing}, 39\penalty0 (1):\penalty0 302--322,
  2009.

\bibitem[Alvarez \& Salzmann(2017)Alvarez and Salzmann]{alvarez2017compression}
Jose~M Alvarez and Mathieu Salzmann.
\newblock Compression-aware training of deep networks.
\newblock In \emph{Advances in Neural Information Processing Systems}, pp.\
  856--867, 2017.

\bibitem[Ardakani et~al.(2016)Ardakani, Condo, and Gross]{ardakani2016sparsely}
Arash Ardakani, Carlo Condo, and Warren~J Gross.
\newblock Sparsely-connected neural networks: towards efficient vlsi
  implementation of deep neural networks.
\newblock \emph{arXiv preprint arXiv:1611.01427}, 2016.

\bibitem[Bingham \& Mannila(2001)Bingham and Mannila]{bingham2001random}
Ella Bingham and Heikki Mannila.
\newblock Random projection in dimensionality reduction: applications to image
  and text data.
\newblock In \emph{Proceedings of the seventh ACM SIGKDD international
  conference on Knowledge discovery and data mining}, pp.\  245--250. ACM,
  2001.

\bibitem[Chin et~al.(2018)Chin, Zhang, and Marculescu]{chin2018layer}
Ting-Wu Chin, Cha Zhang, and Diana Marculescu.
\newblock Layer-compensated pruning for resource-constrained convolutional
  neural networks.
\newblock \emph{arXiv preprint arXiv:1810.00518}, 2018.

\bibitem[Courbariaux et~al.(2016)Courbariaux, Hubara, Soudry, El-Yaniv, and
  Bengio]{courbariaux2016binarized}
Matthieu Courbariaux, Itay Hubara, Daniel Soudry, Ran El-Yaniv, and Yoshua
  Bengio.
\newblock Binarized neural networks: Training deep neural networks with weights
  and activations constrained to+ 1 or-1.
\newblock \emph{arXiv preprint arXiv:1602.02830}, 2016.

\bibitem[Deng et~al.(2009)Deng, Dong, Socher, Li, Li, and
  Fei-Fei]{deng2009imagenet}
Jia Deng, Wei Dong, Richard Socher, Li-Jia Li, Kai Li, and Li~Fei-Fei.
\newblock Imagenet: A large-scale hierarchical image database.
\newblock In \emph{Computer Vision and Pattern Recognition, 2009. CVPR 2009.
  IEEE Conference on}, pp.\  248--255. IEEE, 2009.

\bibitem[Deng et~al.(2018)Deng, Jiao, Pei, Wu, and Li]{deng2018gxnor}
Lei Deng, Peng Jiao, Jing Pei, Zhenzhi Wu, and Guoqi Li.
\newblock Gxnor-net: Training deep neural networks with ternary weights and
  activations without full-precision memory under a unified discretization
  framework.
\newblock \emph{Neural Networks}, 100:\penalty0 49--58, 2018.

\bibitem[Garipov et~al.(2016)Garipov, Podoprikhin, Novikov, and
  Vetrov]{garipov2016ultimate}
Timur Garipov, Dmitry Podoprikhin, Alexander Novikov, and Dmitry Vetrov.
\newblock Ultimate tensorization: compressing convolutional and fc layers
  alike.
\newblock \emph{arXiv preprint arXiv:1611.03214}, 2016.

\bibitem[Goyal et~al.(2017)Goyal, Doll{\'a}r, Girshick, Noordhuis, Wesolowski,
  Kyrola, Tulloch, Jia, and He]{goyal2017accurate}
Priya Goyal, Piotr Doll{\'a}r, Ross Girshick, Pieter Noordhuis, Lukasz
  Wesolowski, Aapo Kyrola, Andrew Tulloch, Yangqing Jia, and Kaiming He.
\newblock Accurate, large minibatch sgd: training imagenet in 1 hour.
\newblock \emph{arXiv preprint arXiv:1706.02677}, 2017.

\bibitem[Han et~al.(2015{\natexlab{a}})Han, Mao, and Dally]{han2015deep}
Song Han, Huizi Mao, and William~J Dally.
\newblock Deep compression: Compressing deep neural networks with pruning,
  trained quantization and huffman coding.
\newblock \emph{arXiv preprint arXiv:1510.00149}, 2015{\natexlab{a}}.

\bibitem[Han et~al.(2015{\natexlab{b}})Han, Pool, Tran, and
  Dally]{han2015learning}
Song Han, Jeff Pool, John Tran, and William Dally.
\newblock Learning both weights and connections for efficient neural network.
\newblock In \emph{Advances in neural information processing systems}, pp.\
  1135--1143, 2015{\natexlab{b}}.

\bibitem[Han et~al.(2016)Han, Liu, Mao, Pu, Pedram, Horowitz, and
  Dally]{han2016eie}
Song Han, Xingyu Liu, Huizi Mao, Jing Pu, Ardavan Pedram, Mark~A Horowitz, and
  William~J Dally.
\newblock Eie: efficient inference engine on compressed deep neural network.
\newblock In \emph{Computer Architecture (ISCA), 2016 ACM/IEEE 43rd Annual
  International Symposium on}, pp.\  243--254. IEEE, 2016.

\bibitem[Han et~al.(2017)Han, Kang, Mao, Hu, Li, Li, Xie, Luo, Yao, Wang,
  et~al.]{han2017ese}
Song Han, Junlong Kang, Huizi Mao, Yiming Hu, Xin Li, Yubin Li, Dongliang Xie,
  Hong Luo, Song Yao, Yu~Wang, et~al.
\newblock Ese: Efficient speech recognition engine with sparse lstm on fpga.
\newblock In \emph{Proceedings of the 2017 ACM/SIGDA International Symposium on
  Field-Programmable Gate Arrays}, pp.\  75--84. ACM, 2017.

\bibitem[He et~al.(2016)He, Zhang, Ren, and Sun]{he2016deep}
Kaiming He, Xiangyu Zhang, Shaoqing Ren, and Jian Sun.
\newblock Deep residual learning for image recognition.
\newblock In \emph{Proceedings of the IEEE conference on computer vision and
  pattern recognition}, pp.\  770--778, 2016.

\bibitem[He et~al.(2018{\natexlab{a}})He, Kang, Dong, Fu, and Yang]{he2018soft}
Yang He, Guoliang Kang, Xuanyi Dong, Yanwei Fu, and Yi~Yang.
\newblock Soft filter pruning for accelerating deep convolutional neural
  networks.
\newblock \emph{arXiv preprint arXiv:1808.06866}, 2018{\natexlab{a}}.

\bibitem[He et~al.(2017)He, Zhang, and Sun]{he2017channel}
Yihui He, Xiangyu Zhang, and Jian Sun.
\newblock Channel pruning for accelerating very deep neural networks.
\newblock In \emph{International Conference on Computer Vision (ICCV)},
  volume~2, pp.\ ~6, 2017.

\bibitem[He et~al.(2018{\natexlab{b}})He, Lin, Liu, Wang, Li, and
  Han]{he2018amc}
Yihui He, Ji~Lin, Zhijian Liu, Hanrui Wang, Li-Jia Li, and Song Han.
\newblock Amc: Automl for model compression and acceleration on mobile devices.
\newblock In \emph{Proceedings of the European Conference on Computer Vision
  (ECCV)}, pp.\  784--800, 2018{\natexlab{b}}.

\bibitem[Hu et~al.(2018)Hu, Sun, Li, Wang, and Gu]{hu2018novel}
Yiming Hu, Siyang Sun, Jianquan Li, Xingang Wang, and Qingyi Gu.
\newblock A novel channel pruning method for deep neural network compression.
\newblock \emph{arXiv preprint arXiv:1805.11394}, 2018.

\bibitem[Ioffe \& Szegedy(2015)Ioffe and Szegedy]{ioffe2015batch}
Sergey Ioffe and Christian Szegedy.
\newblock Batch normalization: Accelerating deep network training by reducing
  internal covariate shift.
\newblock \emph{arXiv preprint arXiv:1502.03167}, 2015.

\bibitem[Jain et~al.(2018)Jain, Phanishayee, Mars, Tang, and
  Pekhimenko]{jain2018gist}
Animesh Jain, Amar Phanishayee, Jason Mars, Lingjia Tang, and Gennady
  Pekhimenko.
\newblock Gist: Efficient data encoding for deep neural network training.
\newblock In \emph{2018 ACM/IEEE 45th Annual International Symposium on
  Computer Architecture (ISCA)}, pp.\  776--789. IEEE, 2018.

\bibitem[Johnson \& Lindenstrauss(1984)Johnson and
  Lindenstrauss]{johnson1984extensions}
William~B Johnson and Joram Lindenstrauss.
\newblock Extensions of lipschitz mappings into a hilbert space.
\newblock \emph{Contemporary mathematics}, 26\penalty0 (189-206):\penalty0 1,
  1984.

\bibitem[Kakade \& Shakhnarovich(2009)Kakade and
  Shakhnarovich]{Kakade_cmsc35900}
Instructors~Sham Kakade and Greg Shakhnarovich.
\newblock Cmsc 35900 (spring 2009) large scale learning lecture: 2 random
  projections, 2009.

\bibitem[Krizhevsky \& Hinton(2009)Krizhevsky and
  Hinton]{krizhevsky2009learning}
Alex Krizhevsky and Geoffrey Hinton.
\newblock Learning multiple layers of features from tiny images.
\newblock 2009.

\bibitem[Krizhevsky et~al.(2012)Krizhevsky, Sutskever, and
  Hinton]{krizhevsky2012imagenet}
Alex Krizhevsky, Ilya Sutskever, and Geoffrey~E Hinton.
\newblock Imagenet classification with deep convolutional neural networks.
\newblock In \emph{Advances in neural information processing systems}, pp.\
  1097--1105, 2012.

\bibitem[LeCun et~al.(1998)LeCun, Bottou, Bengio, and
  Haffner]{lecun1998gradient}
Yann LeCun, L{\'e}on Bottou, Yoshua Bengio, and Patrick Haffner.
\newblock Gradient-based learning applied to document recognition.
\newblock \emph{Proceedings of the IEEE}, 86\penalty0 (11):\penalty0
  2278--2324, 1998.

\bibitem[LeCun et~al.(2015)LeCun, Bengio, and Hinton]{lecun2015deep}
Yann LeCun, Yoshua Bengio, and Geoffrey Hinton.
\newblock Deep learning.
\newblock \emph{nature}, 521\penalty0 (7553):\penalty0 436, 2015.

\bibitem[Leng et~al.(2017)Leng, Li, Zhu, and Jin]{leng2017extremely}
Cong Leng, Hao Li, Shenghuo Zhu, and Rong Jin.
\newblock Extremely low bit neural network: Squeeze the last bit out with admm.
\newblock \emph{arXiv preprint arXiv:1707.09870}, 2017.

\bibitem[Li et~al.(2016)Li, Kadav, Durdanovic, Samet, and Graf]{li2016pruning}
Hao Li, Asim Kadav, Igor Durdanovic, Hanan Samet, and Hans~Peter Graf.
\newblock Pruning filters for efficient convnets.
\newblock \emph{arXiv preprint arXiv:1608.08710}, 2016.

\bibitem[Li et~al.(2006)Li, Hastie, and Church]{li2006very}
Ping Li, Trevor~J Hastie, and Kenneth~W Church.
\newblock Very sparse random projections.
\newblock In \emph{Proceedings of the 12th ACM SIGKDD international conference
  on Knowledge discovery and data mining}, pp.\  287--296. ACM, 2006.

\bibitem[Liang et~al.(2018)Liang, Deng, Zeng, Hu, Ji, Ma, Li, and
  Xie]{liang2018crossbar}
Ling Liang, Lei Deng, Yueling Zeng, Xing Hu, Yu~Ji, Xin Ma, Guoqi Li, and Yuan
  Xie.
\newblock Crossbar-aware neural network pruning.
\newblock \emph{IEEE Access}, 6:\penalty0 58324--58337, 2018.

\bibitem[Lin et~al.(2017{\natexlab{a}})Lin, Sakr, Kim, and
  Shanbhag]{lin2017predictivenet}
Yingyan Lin, Charbel Sakr, Yongjune Kim, and Naresh Shanbhag.
\newblock Predictivenet: An energy-efficient convolutional neural network via
  zero prediction.
\newblock In \emph{Circuits and Systems (ISCAS), 2017 IEEE International
  Symposium on}, pp.\  1--4. IEEE, 2017{\natexlab{a}}.

\bibitem[Lin et~al.(2017{\natexlab{b}})Lin, Han, Mao, Wang, and
  Dally]{lin2017deep}
Yujun Lin, Song Han, Huizi Mao, Yu~Wang, and William~J Dally.
\newblock Deep gradient compression: Reducing the communication bandwidth for
  distributed training.
\newblock \emph{arXiv preprint arXiv:1712.01887}, 2017{\natexlab{b}}.

\bibitem[Liu et~al.(2017)Liu, Li, Shen, Huang, Yan, and Zhang]{liu2017learning}
Zhuang Liu, Jianguo Li, Zhiqiang Shen, Gao Huang, Shoumeng Yan, and Changshui
  Zhang.
\newblock Learning efficient convolutional networks through network slimming.
\newblock In \emph{2017 IEEE International Conference on Computer Vision
  (ICCV)}, pp.\  2755--2763. IEEE, 2017.

\bibitem[Luo \& Wu(2018)Luo and Wu]{luo2018autopruner}
Jian-Hao Luo and Jianxin Wu.
\newblock Autopruner: An end-to-end trainable filter pruning method for
  efficient deep model inference.
\newblock \emph{arXiv preprint arXiv:1805.08941}, 2018.

\bibitem[Luo et~al.(2017)Luo, Wu, and Lin]{luo2017thinet}
Jian-Hao Luo, Jianxin Wu, and Weiyao Lin.
\newblock Thinet: A filter level pruning method for deep neural network
  compression.
\newblock \emph{arXiv preprint arXiv:1707.06342}, 2017.

\bibitem[McKinstry et~al.(2018)McKinstry, Esser, Appuswamy, Bablani, Arthur,
  Yildiz, and Modha]{mckinstry2018discovering}
Jeffrey~L McKinstry, Steven~K Esser, Rathinakumar Appuswamy, Deepika Bablani,
  John~V Arthur, Izzet~B Yildiz, and Dharmendra~S Modha.
\newblock Discovering low-precision networks close to full-precision networks
  for efficient embedded inference.
\newblock \emph{arXiv preprint arXiv:1809.04191}, 2018.

\bibitem[Mishkin \& Matas(2015)Mishkin and Matas]{mishkin2015all}
Dmytro Mishkin and Jiri Matas.
\newblock All you need is a good init.
\newblock \emph{arXiv preprint arXiv:1511.06422}, 2015.

\bibitem[Molchanov et~al.(2016)Molchanov, Tyree, Karras, Aila, and
  Kautz]{molchanov2016pruning}
Pavlo Molchanov, Stephen Tyree, Tero Karras, Timo Aila, and Jan Kautz.
\newblock Pruning convolutional neural networks for resource efficient
  inference.
\newblock 2016.

\bibitem[Morcos et~al.(2018)Morcos, Barrett, Rabinowitz, and
  Botvinick]{morcos2018importance}
Ari~S Morcos, David~GT Barrett, Neil~C Rabinowitz, and Matthew Botvinick.
\newblock On the importance of single directions for generalization.
\newblock \emph{arXiv preprint arXiv:1803.06959}, 2018.

\bibitem[Novikov et~al.(2015)Novikov, Podoprikhin, Osokin, and
  Vetrov]{novikov2015tensorizing}
Alexander Novikov, Dmitrii Podoprikhin, Anton Osokin, and Dmitry~P Vetrov.
\newblock Tensorizing neural networks.
\newblock In \emph{Advances in Neural Information Processing Systems}, pp.\
  442--450, 2015.

\bibitem[Redmon \& Farhadi(2016)Redmon and Farhadi]{redmon2016yolo9000}
Joseph Redmon and Ali Farhadi.
\newblock Yolo9000: better, faster, stronger.
\newblock \emph{arXiv preprint}, 1612, 2016.

\bibitem[Rhu et~al.(2018)Rhu, O'Connor, Chatterjee, Pool, Kwon, and
  Keckler]{rhu2018compressing}
Minsoo Rhu, Mike O'Connor, Niladrish Chatterjee, Jeff Pool, Youngeun Kwon, and
  Stephen~W Keckler.
\newblock Compressing dma engine: Leveraging activation sparsity for training
  deep neural networks.
\newblock In \emph{High Performance Computer Architecture (HPCA), 2018 IEEE
  International Symposium on}, pp.\  78--91. IEEE, 2018.

\bibitem[Simonyan \& Zisserman(2014)Simonyan and Zisserman]{simonyan2014very}
Karen Simonyan and Andrew Zisserman.
\newblock Very deep convolutional networks for large-scale image recognition.
\newblock \emph{arXiv preprint arXiv:1409.1556}, 2014.

\bibitem[Smith et~al.(2017)Smith, Kindermans, and Le]{smith2017don}
Samuel~L Smith, Pieter-Jan Kindermans, and Quoc~V Le.
\newblock Don't decay the learning rate, increase the batch size.
\newblock \emph{arXiv preprint arXiv:1711.00489}, 2017.

\bibitem[Spring \& Shrivastava(2017)Spring and Shrivastava]{spring2017scalable}
Ryan Spring and Anshumali Shrivastava.
\newblock Scalable and sustainable deep learning via randomized hashing.
\newblock In \emph{Proceedings of the 23rd ACM SIGKDD International Conference
  on Knowledge Discovery and Data Mining}, pp.\  445--454. ACM, 2017.

\bibitem[Sun et~al.(2017)Sun, Ren, Ma, and Wang]{sun2017meprop}
Xu~Sun, Xuancheng Ren, Shuming Ma, and Houfeng Wang.
\newblock meprop: Sparsified back propagation for accelerated deep learning
  with reduced overfitting.
\newblock \emph{arXiv preprint arXiv:1706.06197}, 2017.

\bibitem[Vijaykumar et~al.(2015)Vijaykumar, Pekhimenko, Jog, Bhowmick,
  Ausavarungnirun, Das, Kandemir, Mowry, and Mutlu]{vijaykumar2015case}
Nandita Vijaykumar, Gennady Pekhimenko, Adwait Jog, Abhishek Bhowmick, Rachata
  Ausavarungnirun, Chita Das, Mahmut Kandemir, Todd~C Mowry, and Onur Mutlu.
\newblock A case for core-assisted bottleneck acceleration in gpus: enabling
  flexible data compression with assist warps.
\newblock In \emph{ACM SIGARCH Computer Architecture News}, volume~43, pp.\
  41--53. ACM, 2015.

\bibitem[Vu(2016)]{vu2016random}
Khac~Ky Vu.
\newblock \emph{Random projection for high-dimensional optimization}.
\newblock PhD thesis, Universit{\'e} Paris-Saclay, 2016.

\bibitem[Wang et~al.(2014)Wang, Zhang, Shen, Zhang, Lu, Wu, and
  Wang]{wang2014intel}
Endong Wang, Qing Zhang, Bo~Shen, Guangyong Zhang, Xiaowei Lu, Qing Wu, and
  Yajuan Wang.
\newblock Intel math kernel library.
\newblock In \emph{High-Performance Computing on the Intel{\textregistered}
  Xeon Phi™}, pp.\  167--188. Springer, 2014.

\bibitem[Wen et~al.(2016)Wen, Wu, Wang, Chen, and Li]{wen2016learning}
Wei Wen, Chunpeng Wu, Yandan Wang, Yiran Chen, and Hai Li.
\newblock Learning structured sparsity in deep neural networks.
\newblock In \emph{Advances in Neural Information Processing Systems}, pp.\
  2074--2082, 2016.

\bibitem[Wen et~al.(2017)Wen, Xu, Yan, Wu, Wang, Chen, and Li]{wen2017terngrad}
Wei Wen, Cong Xu, Feng Yan, Chunpeng Wu, Yandan Wang, Yiran Chen, and Hai Li.
\newblock Terngrad: Ternary gradients to reduce communication in distributed
  deep learning.
\newblock In \emph{Advances in Neural Information Processing Systems}, pp.\
  1508--1518, 2017.

\bibitem[Wu et~al.(2018)Wu, Li, Chen, and Shi]{wu2018training}
Shuang Wu, Guoqi Li, Feng Chen, and Luping Shi.
\newblock Training and inference with integers in deep neural networks.
\newblock \emph{arXiv preprint arXiv:1802.04680}, 2018.

\bibitem[Wu et~al.(2016)Wu, Schuster, Chen, Le, Norouzi, Macherey, Krikun, Cao,
  Gao, Macherey, et~al.]{wu2016google}
Yonghui Wu, Mike Schuster, Zhifeng Chen, Quoc~V Le, Mohammad Norouzi, Wolfgang
  Macherey, Maxim Krikun, Yuan Cao, Qin Gao, Klaus Macherey, et~al.
\newblock Google's neural machine translation system: Bridging the gap between
  human and machine translation.
\newblock \emph{arXiv preprint arXiv:1609.08144}, 2016.

\bibitem[Wu \& He(2018)Wu and He]{wu2018group}
Yuxin Wu and Kaiming He.
\newblock Group normalization.
\newblock \emph{arXiv preprint arXiv:1803.08494}, 2018.

\bibitem[Xiao et~al.(2017)Xiao, Rasul, and Vollgraf]{xiao2017fashion}
Han Xiao, Kashif Rasul, and Roland Vollgraf.
\newblock Fashion-mnist: a novel image dataset for benchmarking machine
  learning algorithms.
\newblock \emph{arXiv preprint arXiv:1708.07747}, 2017.

\bibitem[Xue et~al.(2014)Xue, Li, Yu, Seltzer, and Gong]{xue2014singular}
Jian Xue, Jinyu Li, Dong Yu, Mike Seltzer, and Yifan Gong.
\newblock Singular value decomposition based low-footprint speaker adaptation
  and personalization for deep neural network.
\newblock In \emph{Acoustics, Speech and Signal Processing (ICASSP), 2014 IEEE
  International Conference on}, pp.\  6359--6363. IEEE, 2014.

\bibitem[Yang et~al.(2017)Yang, Krompass, and Tresp]{yang2017tensor}
Yinchong Yang, Denis Krompass, and Volker Tresp.
\newblock Tensor-train recurrent neural networks for video classification.
\newblock \emph{arXiv preprint arXiv:1707.01786}, 2017.

\bibitem[Ye et~al.(2018)Ye, Lu, Lin, and Wang]{ye2018rethinking}
Jianbo Ye, Xin Lu, Zhe Lin, and James~Z Wang.
\newblock Rethinking the smaller-norm-less-informative assumption in channel
  pruning of convolution layers.
\newblock \emph{arXiv preprint arXiv:1802.00124}, 2018.

\bibitem[You et~al.(2017)You, Zhang, Hsieh, Demmel, and
  Keutzer]{you2017imagenet}
Yang You, Zhao Zhang, C~Hsieh, James Demmel, and Kurt Keutzer.
\newblock Imagenet training in minutes.
\newblock \emph{CoRR, abs/1709.05011}, 2017.

\bibitem[Zagoruyko \& Komodakis(2016)Zagoruyko and
  Komodakis]{zagoruyko2016wide}
Sergey Zagoruyko and Nikos Komodakis.
\newblock Wide residual networks.
\newblock \emph{arXiv preprint arXiv:1605.07146}, 2016.

\bibitem[Zhang et~al.(2018)Zhang, Zhang, Ye, Li, Tang, Wen, Lin, Fardad, and
  Wang]{zhang2018adam}
Tianyun Zhang, Kaiqi Zhang, Shaokai Ye, Jiayu Li, Jian Tang, Wujie Wen, Xue
  Lin, Makan Fardad, and Yanzhi Wang.
\newblock Adam-admm: A unified, systematic framework of structured weight
  pruning for dnns.
\newblock \emph{arXiv preprint arXiv:1807.11091}, 2018.

\bibitem[Zhang et~al.(2000)Zhang, Yang, and Gupta]{zhang2000frequent}
Youtao Zhang, Jun Yang, and Rajiv Gupta.
\newblock Frequent value locality and value-centric data cache design.
\newblock \emph{ACM SIGPLAN Notices}, 35\penalty0 (11):\penalty0 150--159,
  2000.

\bibitem[Zhou et~al.(2016)Zhou, Wu, Ni, Zhou, Wen, and Zou]{zhou2016dorefa}
Shuchang Zhou, Yuxin Wu, Zekun Ni, Xinyu Zhou, He~Wen, and Yuheng Zou.
\newblock Dorefa-net: Training low bitwidth convolutional neural networks with
  low bitwidth gradients.
\newblock \emph{arXiv preprint arXiv:1606.06160}, 2016.

\end{thebibliography}
\bibliographystyle{iclr2019_conference}

\newpage
\begin{appendices}

\section{Proof of the Dimension-reduction Search for Inner Product Preservation}\label{app:proof}

\textbf{Theorem 1.} Given a set of $N$ points in $\mathbb{R}^d$ (i.e. all $\textbf{X}_i$ and $\textbf{W}_j$), and a number of $k>O(\frac{log(N)}{\epsilon ^2})$, there exist a linear map $f: \mathbb{R}^d\Rightarrow \mathbb{R}^k$ and a $\epsilon_0 \in (0, 1)$, for $0<\epsilon \leq \epsilon_0$ we have 
\begin{equation}
P[~| \langle f(\textbf{X}_i), f(\textbf{W}_j) \rangle - \langle \textbf{X}_i, \textbf{W}_j \rangle |\leq \epsilon~]\geq 1-O(\epsilon ^2).
\label{Proof0}
\end{equation}
for all $\textbf{X}_i$ and $\textbf{W}_j$.

\textbf{Proof}. According to the definition of inner product and vector norm, any two vectors $\textbf{a}$ and $\textbf{b}$ satisfy
\begin{equation}
\begin{cases}
\langle \textbf{a}, \textbf{b} \rangle = (\Vert \textbf{a} \Vert ^2 + \Vert \textbf{b} \Vert ^2 - \Vert \textbf{a} - \textbf{b} \Vert ^2)/2 \\
\langle \textbf{a}, \textbf{b} \rangle = (\Vert \textbf{a} + \textbf{b} \Vert ^2 - \Vert \textbf{a} \Vert ^2 - \Vert \textbf{b} \Vert ^2)/2 \\
\end{cases}.
\label{Inner_definition1}
\end{equation}
It is easy to further get 
\begin{equation}
\langle \textbf{a}, \textbf{b} \rangle = (\Vert \textbf{a} + \textbf{b} \Vert ^2 - \Vert \textbf{a} - \textbf{b} \Vert ^2)/4.
\label{Inner_definition2}
\end{equation}
Therefore, we can transform the target in equation (\ref{Proof0}) to
\begin{equation}
\begin{array}{c}
|~\langle f(\textbf{X}_i), f(\textbf{W}_j) \rangle - \langle \textbf{X}_i, \textbf{W}_j \rangle~| 
\\ = |~\Vert f(\textbf{X}_i) + f(\textbf{W}_j) \Vert ^2 - \Vert f(\textbf{X}_i) - f(\textbf{W}_j) \Vert ^2 - \Vert \textbf{X}_i + \textbf{W}_j \Vert ^2 + \Vert \textbf{X}_i - \textbf{W}_j \Vert ^2~|/4
\\ \leq |~\Vert f(\textbf{X}_i) + f(\textbf{W}_j) \Vert ^2 - \Vert \textbf{X}_i + \textbf{W}_j \Vert ^2~|/4 + |~\Vert f(\textbf{X}_i) - f(\textbf{W}_j) \Vert ^2 - \Vert \textbf{X}_i - \textbf{W}_j \Vert ^2~|/4,
\end{array}
\label{Proof1}
\end{equation}
which is also based on the fact that $|u-v|\leq|u|+|v|$. Now recall the definition of random projection in equation (5) of the main text
\begin{equation}
f(\textbf{X}_i) = \frac{1}{\sqrt{k}}\textbf{R}\textbf{X}_i\in \mathbb{R}^k,~~f(\textbf{W}_j) = \frac{1}{\sqrt{k}}\textbf{R}\textbf{W}_j\in \mathbb{R}^k.
\label{RP}
\end{equation}
Substituting equation (\ref{RP}) into equation (\ref{Proof1}), we have
\begin{equation}
\begin{array}{c}
|~\langle f(\textbf{X}_i), f(\textbf{W}_j) \rangle - \langle \textbf{X}_i, \textbf{W}_j \rangle~| 
\\ \leq |~\Vert \frac{1}{\sqrt{k}}\textbf{R}\textbf{X}_i + \frac{1}{\sqrt{k}}\textbf{R}\textbf{W}_j \Vert ^2 - \Vert \textbf{X}_i + \textbf{W}_j \Vert ^2~|/4 + |~\Vert \frac{1}{\sqrt{k}}\textbf{R}\textbf{X}_i - \frac{1}{\sqrt{k}}\textbf{R}\textbf{W}_j \Vert ^2 - \Vert \textbf{X}_i - \textbf{W}_j \Vert ^2~|/4
\\ = |~\Vert \frac{1}{\sqrt{k}}\textbf{R}(\textbf{X}_i + \textbf{W}_j) \Vert ^2 - \Vert \textbf{X}_i + \textbf{W}_j \Vert ^2~|/4 + |~\Vert \frac{1}{\sqrt{k}}\textbf{R}(\textbf{X}_i - \textbf{W}_j) \Vert ^2 - \Vert \textbf{X}_i - \textbf{W}_j \Vert ^2~|/4
\\ = |~\Vert f(\textbf{X}_i + \textbf{W}_j) \Vert ^2 - \Vert \textbf{X}_i + \textbf{W}_j \Vert ^2~|/4 + |~\Vert f(\textbf{X}_i - \textbf{W}_j) \Vert ^2 - \Vert \textbf{X}_i - \textbf{W}_j \Vert ^2~|/4
\end{array}.
\label{Proof2}
\end{equation}
Further recalling the norm preservation in equation (3) of the main text: there exist a linear map $f: \mathbb{R}^d\Rightarrow \mathbb{R}^k$ and a $\epsilon_0 \in (0, 1)$, for $0<\epsilon \leq \epsilon_0$ we have
\begin{equation}
P[~(1-\epsilon)\Vert \textbf{Z}\Vert ^2 \leq \Vert f(\textbf{Z})\Vert ^2 \leq (1+\epsilon)\Vert \textbf{Z}\Vert ^2~]\geq 1-O(\epsilon ^2).
\label{JL_norm}
\end{equation}
Substituting the equation (\ref{JL_norm}) into equation (\ref{Proof2}) yields
\begin{equation}
\begin{array}{c}
P[~|~\Vert f(\textbf{X}_i + \textbf{W}_j) \Vert ^2 - \Vert \textbf{X}_i + \textbf{W}_j \Vert ^2~|/4 + |~\Vert f(\textbf{X}_i - \textbf{W}_j) \Vert ^2 - \Vert \textbf{X}_i - \textbf{W}_j \Vert ^2~|/4... 
\\ \leq \frac{\epsilon}{4}(\Vert \textbf{X}_i + \textbf{W}_j \Vert ^2 + \Vert \textbf{X}_i - \textbf{W}_j \Vert ^2)=\frac{\epsilon}{2}(\Vert \textbf{X}_i \Vert ^2 + \Vert \textbf{W}_j \Vert ^2)~]...
\\ \geq P(~|~\Vert f(\textbf{X}_i + \textbf{W}_j) \Vert ^2 - \Vert \textbf{X}_i + \textbf{W}_j \Vert ^2~|/4\leq \frac{\epsilon}{4}\Vert \textbf{X}_i + \textbf{W}_j \Vert ^2~)...
\\ \times P(~|~\Vert f(\textbf{X}_i - \textbf{W}_j) \Vert ^2 - \Vert \textbf{X}_i - \textbf{W}_j \Vert ^2~|/4\leq \frac{\epsilon}{4}\Vert \textbf{X}_i - \textbf{W}_j \Vert ^2~)...
\\ \geq [1-O(\epsilon ^2)]\cdot [1-O(\epsilon ^2)]=1-O(\epsilon ^2).
\end{array}.
\label{Proof3}
\end{equation}
Combining equation (\ref{Proof2}) and (\ref{Proof3}), finally we have
\begin{equation}
\begin{array}{c}
P[~|~\langle f(\textbf{X}_i), f(\textbf{W}_j) \rangle - \langle \textbf{X}_i, \textbf{W}_j \rangle~| \leq \frac{\epsilon}{2}(\Vert \textbf{X}_i \Vert ^2 + \Vert \textbf{W}_j \Vert ^2)~] \geq 1-O(\epsilon ^2)
\end{array}.
\label{Proof4}
\end{equation}

It can be seen that, for any given $\textbf{X}_i$ and $\textbf{W}_j$ pair, the inner product can be preserved if the $\epsilon$ is sufficiently small. Actually, previous work \citep{achlioptas2001database, bingham2001random, vu2016random} discussed a lot on the random projection for various big data applications, here we re-organize these supporting materials to form a systematic proof. We hope this could help readers to follow this paper. In practical experiments, there exists a trade-off between the dimension reduction degree and the recognition accuracy. Smaller $\epsilon$ usually brings more accurate inner product estimation and better recognition accuracy while at the cost of higher computational complexity with larger $k$, and vice versa. Because the $\Vert \textbf{X}_i \Vert ^2$ and $\Vert \textbf{W}_j \Vert ^2$ are not strictly bounded, the approximation may suffer from some noises. Anyway, from the abundant experiments in this work, the effectiveness of our  approach for training dynamic and sparse neural networks has been validated.

\section{Implementation and overhead}\label{app:DRS}

\begin{algorithm}[!htbp]
\vspace{-10pt}
\caption{DSG training}\label{algorithm} 
\label{Algorithm}
\KwData{A mini-batch of inputs \& targets ($\textbf{X}_0$, $\textbf{X}^*$), previous weights $\textbf{W}^t$, previous BN parameters $\theta^t$.}  
\KwResult{Update weights $\textbf{W}^{t+1}$, update BN parameters $\theta^{t+1}$.}
~\\
Random projection: $f(\textbf{W}_k^t)\Leftarrow \textbf{W}_k^t$\;
~\\
Step 1. Forward Computation\;
\For{k=1 to L}
{
\eIf{k$<$L}
{
Projection: $f(\textbf{X}_{k-1})\Leftarrow \textbf{X}_{k-1}$\;
Generating $Mask_k$ via dimension-reduction search according to $f(\textbf{X}_{k-1})$ and $f(\textbf{W}_k^t)$\;
${\color{red}\textbf{S}_k} \Leftarrow \varphi[~Mask_k({\color{red}\textbf{X}_{k-1}}\textbf{W}_k^t)~]$\;
${\color{red}\textbf{X}_k} \Leftarrow Mask_k[~BN({\color{red}\textbf{S}_k}, \theta_k^t)~]$\;
}
{
$\textbf{X}_L \Leftarrow linear({\color{red}\textbf{X}_{L-1}}\textbf{W}_L^t)$\;
}
}
~\\

Step 2. Backward Computation\;
Compute the gradient of the output layer $\textbf{G}_{\textbf{X}_L}=\frac{\partial C(\textbf{X}_L, \textbf{X}^*)}{\partial \textbf{X}_L}$\;
\For{k=L to 1}
{
\eIf{k==L}
{
${\color{red}\textbf{G}_{\textbf{X}_{L-1}}}\Leftarrow Mask_{k-1}(\textbf{G}_{\textbf{X}_L}(\textbf{W}_L^t)^T)$\;
$\textbf{G}_{\textbf{W}_L}\Leftarrow \textbf{G}_{\textbf{X}_L}^T {\color{red}\textbf{X}_{L-1}}$\;
}
{
$({\color{red}\textbf{G}_{\textbf{S}_k}}, \textbf{G}_{\theta_k})\Leftarrow Mask_k[~BN\_{grad}({\color{red}\textbf{G}_{\textbf{X}_k}}, {\color{red}\textbf{S}_k}, \theta_k^t)~]$\;
$\textbf{G}_{\textbf{W}_k}\Leftarrow ({\color{red} \textbf{G}_{\textbf{S}_k}\odot \varphi \_grad})^T {\color{red}\textbf{X}_{k-1}}$\;
\If{k$>$1}
{${\color{red}\textbf{G}_{\textbf{X}_{k-1}}}\Leftarrow Mask_{k-1}[~({\color{red} \textbf{G}_{\textbf{S}_k}\odot \varphi\_{grad}})(\textbf{W}_k^t)^T~]$\;}
}
}
~\\

Step 3. Parameter Update\;
\For{k=1 to L}
{
$\textbf{W}_k^{t+1}\Leftarrow Optimizer(\textbf{W}_k^t, \textbf{G}_{\textbf{W}_k})$\;
$\theta_k^{t+1}\Leftarrow Optimizer(\theta_k^t, \textbf{G}_{\theta_k})$\;
}
\end{algorithm}

\begin{figure}[!htbp]
\centering
\includegraphics[width=0.7\textwidth]{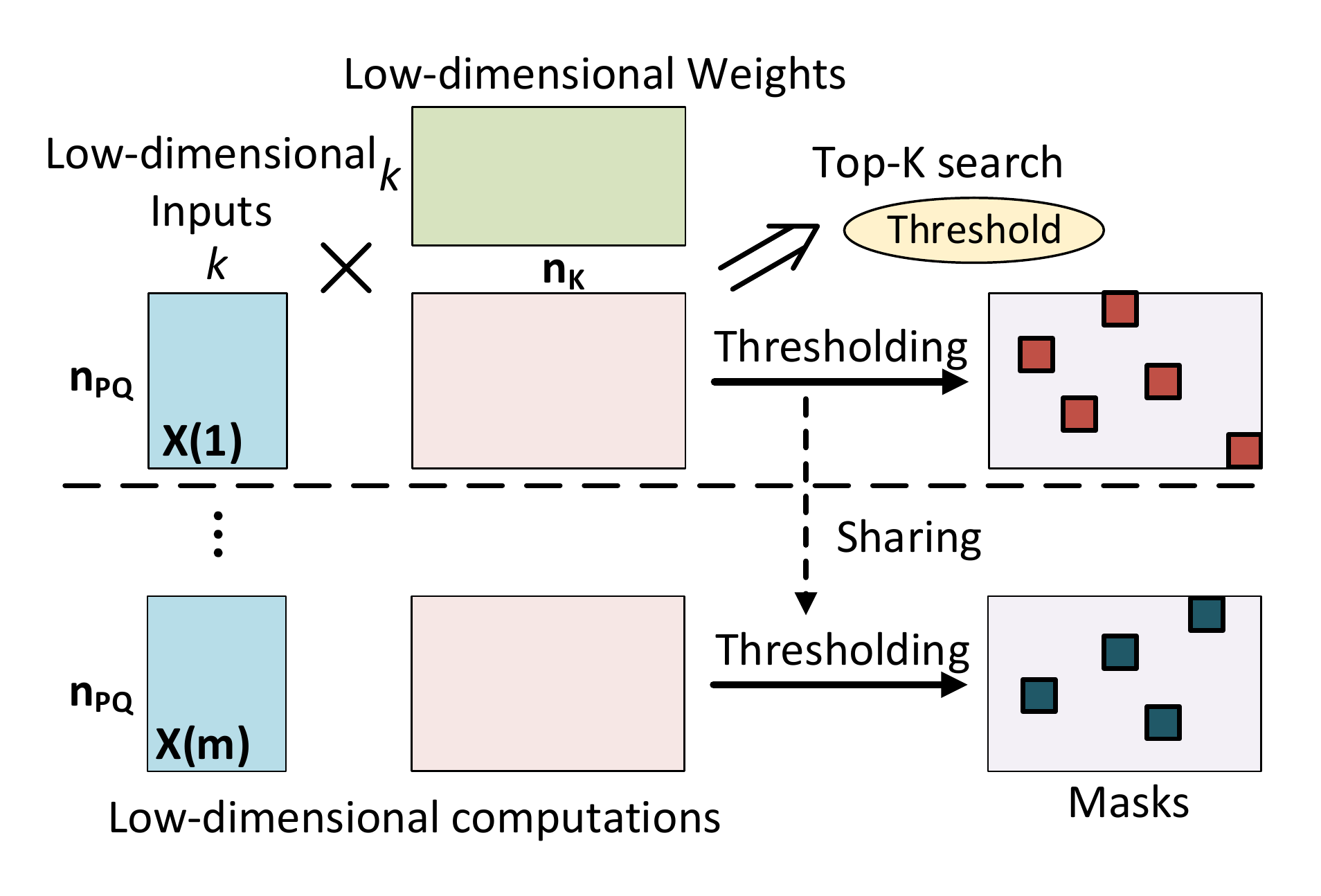}
\vspace{-8pt}
\caption{Selection mask generation: using a top-\textit{k} search on the first input sample $X(1)$ within each mini-batch to obtain a top-\textit{k} threshold which is shared by the following samples. Then, we apply thresholding on the whole output activation tensor to generate the importance mask for the same mini-batch.}
\label{DRS}
\end{figure}

The training algorithm for generating DSG is presented in Algorithm \ref{algorithm}. The generation procedure of the critical neuron mask based on the virtual activations estimated in the low-dimensional space is presented in Figure \ref{DRS}, which is a typical top-\textit{k} search. The \textit{k} value is determined by the activation size and the desired sparsity $\gamma$. To reduce the search cost, we calculate the first input sample $X(1)$ within the current mini-batch and then conduct a top-\textit{k} search over the whole virtual activation matrix for obtaining the top-\textit{k} threshold under this sample. The remaining samples share the top-\textit{k} threshold from the first sample to avoid costly searching overhead. At last, the overall activation mask is generated by setting the mask element to one if the estimated activation is larger than the top-\textit{k} threshold and setting others to zero. In this way, we greatly reduce the search cost. Note that, for the FC layer, each sample $X(i)$ is a vector.

\begin{table*}[!htbp]
	\centering 
	\caption{Computational complexity of dimension-reduction search. MMACs denotes mega-MACs and BL denotes baseline.}
	\label{tab:compute}
    \resizebox{0.95\textwidth}{!}{
	\begin{tabular}{c|lllll|lllll}
		\hline \hline
Layers & \multicolumn{5}{c|}{Dimension} & \multicolumn{5}{c}{Operations (MMACs)} \\ \hline
$n_{PQ}$, $n_{CRS}$, $n_K$ & BL & 0.3 & 0.5 & 0.7 & 0.9 & BL & 0.3 & 0.5 & 0.7 & 0.9 \\ \hline
1024, 1152, 128 & 1152 & 539 & 232 & 148 & 119 & 144 & 67.37 & 29 & 18.5 & 14.88 \\ \hline
256, 1152, 256 & 1152 & 616 & 266 & 169 & 136 & 72 & 38.5 & 16.63 & 10.56 & 8.5 \\ \hline
256, 2304, 256 & 2304 & 616 & 266 & 169 & 136 & 144 & 38.5 & 16.63 & 10.56 & 8.5 \\ \hline
64, 2304, 512 & 2304 & 693 & 299 & 190 & 154 & 72 & 21.65 & 9.34 & 5.94 & 4.81 \\ \hline
64, 4608, 512 & 4608 & 693 & 299 & 190 & 154 & 144 & 21.65 & 9.34 & 5.94 & 4.81 \\ \hline
		\hline
	\end{tabular}}
\end{table*}

Furthermore, we investigate the influence of the $\epsilon$ on the computation cost of dimension-reduction search for importance estimation. We take several layers from the VGG8 on CIFAR10 as a case study, as shown in Table \ref{tab:compute}. With $\epsilon$ larger, the dimension-reduction search can achieve lower dimension with much fewer operations. The average reduction of the dimension is 3.6x ($\epsilon=0.3$), 8.5x ($\epsilon=0.5$), 13.3x ($\epsilon=0.7$), and 16.5x ($\epsilon=0.9$). The resulting operation reduction is 3.1x, 7.1x, 11.1x, and 13.9x, respectively. 

\section{Convergence Analysis}\label{app:convergence}

\begin{figure}[!htbp]
\centering
\includegraphics[width=0.98\textwidth]{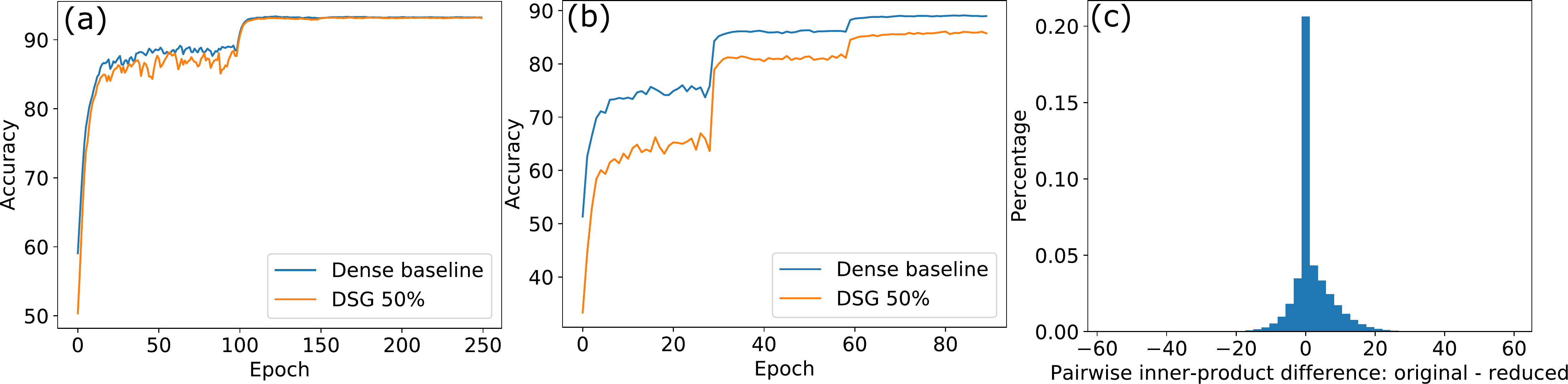}
\vspace{-8pt}
\caption{Accuracy convergence. (a) Training curve with validation accuracy of VGG8 on CIFAR10; (b) Training curve with top-5 validation accuracy of ResNet-18 on ImageNet; (c) Distribution of pairwise difference between the original high-dimensional inner product and the low-dimensional one for the CONV5 layer in VGG8.}
\label{Acc_Convergence}
\end{figure}

One interesting question is that whether DSG slows down the training convergence or not, which is answered by Figure \ref{Acc_Convergence}. According to Figure \ref{Acc_Convergence}(a)-(b), the convergence speed under DSG constraints varies little from the vanilla model training. This probably owes to the high fidelity of inner product when we use random projection to reduce the data dimension. Figure \ref{Acc_Convergence}(c) visualizes the distribution of the pairwise difference between the original high-dimensional inner product and the low-dimensional one for the CONV5 layer of VGG8 on CIFAR10. Most of the inner product differences are around zero, which implies an accurate approximation capability of the proposed dimension-reduction search. This helps reduce the training variance and avoid training deceleration.

\begin{figure}[!htbp]
\centering
\includegraphics[width=0.95\textwidth]{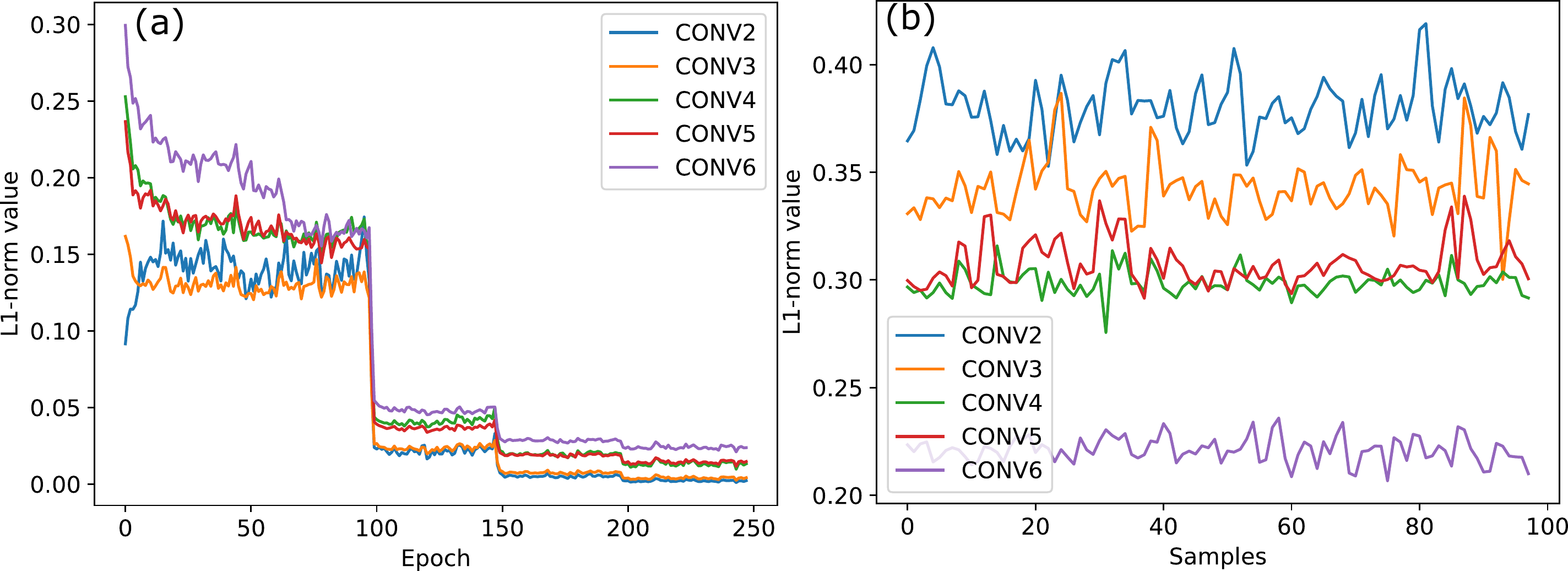}
\vspace{-8pt}
\caption{Selection mask convergence. (a) Average $L1$-norm value of the difference mask tensors between adjacent training epochs across all samples in one mini-batch; (b) Average $L1$-norm value of the difference mask tensors between adjacent samples after training.}
\label{Mask_Convergence}
\end{figure}

Another question in DSG is that whether the selection masks converge during training or not. To explore the answer, we did an additional experiment as shown in the Figure \ref{Mask_Convergence}. We select a mini-batch of training samples as a case study for data recording. Each curve presents the results of one layer (CONV2-CONV6). For each sample at each layer, we recorded the change of binary selection mask between two adjacent training epochs. Here the change is obtained by calculating the $L1$-norm value of the difference tensor of two mask tensors at two adjacent epochs, i.e., $change=batch\_avg\_L1norm(mask_{i+1}-mask_i)$. Here the $batch\_avg\_L1norm(\cdot)$ indicates the average $L1$-norm value across all samples in one mini-batch. As shown in Figure \ref{Mask_Convergence}(a), the selection mask for each sample converges as training goes on.

In our implementation we inherit the random projection matrix from training and perform the same on-the-fly dimension-reduction search in inference. We didn't try to directly suspend the selection masks, because the selection mask might vary across samples even if we observe convergence for each sample. This can be seen from Figure \ref{Mask_Convergence}(b), where the difference mask tensors between adjacent samples in one mini-batch present significant differences (large $L1$-norm value) after training. Therefore, it will consume lot of memory space to save these trained masks for all samples, which is less efficient than conducting on-the-fly search during inference.

\section{Comparison with Other Methods}\label{app:comparison}

\begin{figure}[!htbp]
\centering
\includegraphics[width=0.6\textwidth]{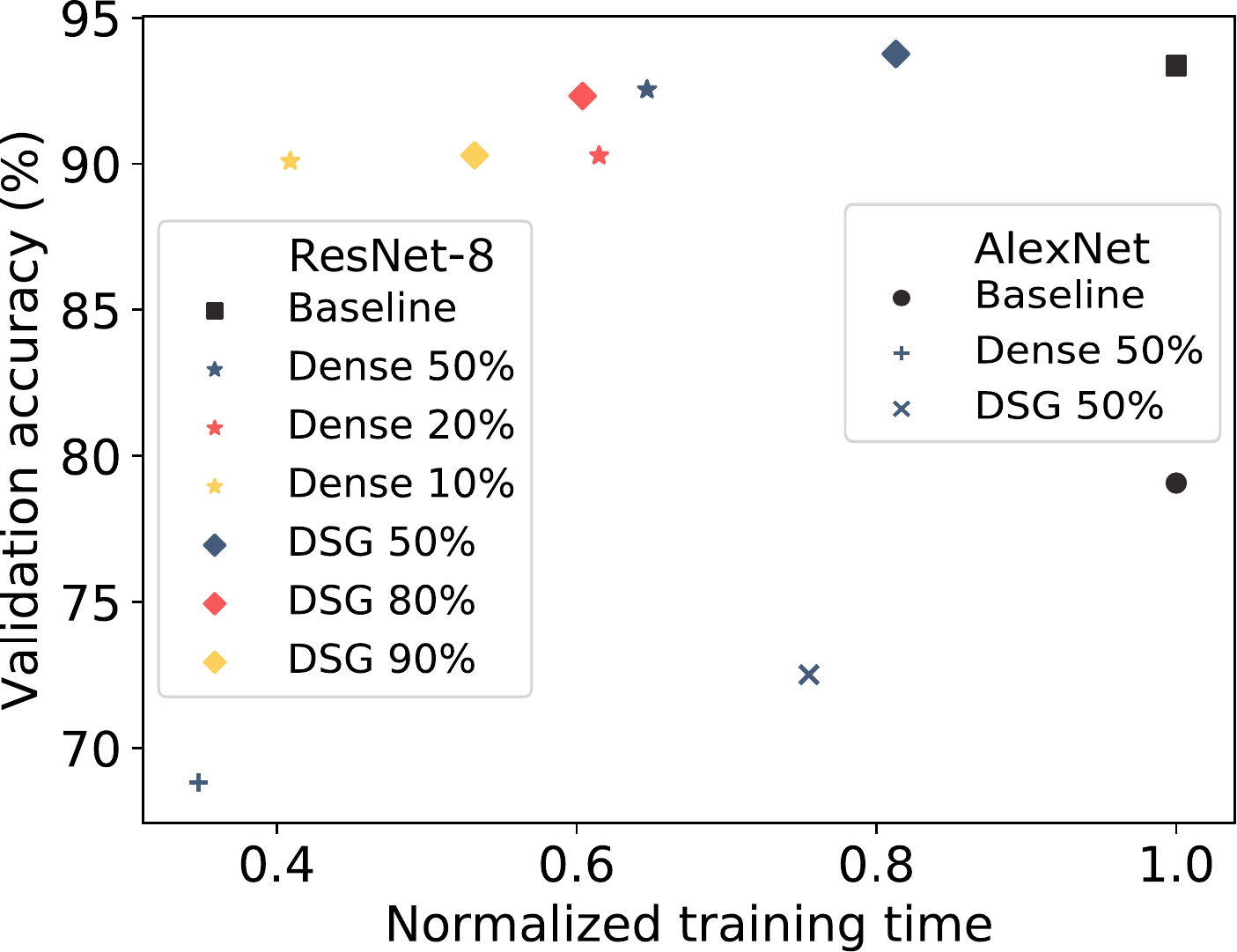}
\vspace{-8pt}
\caption{Comparison with smaller-dense models with equivalent MACs using ResNet8 on CIFAR10 and AlexNet on ImageNet.}
\label{vs_small}
\end{figure}

Figure \ref{vs_small} extends Figure \ref{measured}(b) in the main text to more network structures, including ResNet8 on CIFAR10 and AlexNet on ImageNet. The similar observation can be achieved: the equivalent smaller dense models with the same effective MACs are able to save more training time but the accuracy degradation will be increased. Note that in this figure, the DSG training uses a warm-up training with dense model for the first 10 epochs. The overhead of the warm-up training has been taken account into the entire training cost. To make the accuracy results on CIFAR10 and ImageNet comparable for figure clarity, AlexNet reports the top-5 accuracy.

Our work targets at both the training and inference phases while most of previous work focused on the inference compression. In prior methods, the training usually becomes more complicated with various regularization constraints or iterative fine-tuning/retraining. Therefore, it is not very fair to compare with them during training. For this reason, we just compare with them on the inference pruning. Different from doing DSG training from scratch, here we utilize DSG for fine-tuning based on pre-trained models.

\begin{table*}[!htbp]
	\centering 
	\caption{Comparison with other structured sparsification methods for inference. All the results are from VGG16 on ImageNet, and the default accuracy is top-1 accuracy. The baseline methods are Taylor Expansion \citep{molchanov2016pruning}, ThiNet \citep{luo2017thinet}, Channel Pruning \citep{hu2018novel}, AutoPrunner \citep{luo2018autopruner}, and AMC \citep{he2018amc}.}
	\label{tab:comparison}
	\renewcommand\arraystretch{1.3}
    \resizebox{0.97\textwidth}{!}{
	\begin{tabular}{c|ccccc|c}
		\hline \hline
Methods & Taylor Expansion & ThiNet & Channel Pruning & AutoPrunner & AMC & \textbf{DSG} \\ \hline 
Operation Sparsity & 62.86\% & 69.81\% & 69.32\% & 73.6\% & 80\% & 62.92\% \\ \hline
\multirow{2}*{Accuracy} & \multirow{2}*{87\%(top-5)} & \multirow{2}*{67.34\%} & \multirow{2}*{70.42\%} & \multirow{2}*{68.43\%} & \multirow{2}*{69.1\%} & 71.44\%(top-1) \\ 
 &  &  &  &  &  & 90.56\%(top-5) \\ \hline \hline
	\end{tabular}}
\end{table*}

To guarantee the fairness, all the results are from the same network (VGG16) on the same dataset (ImageNet). Since our DSG produces structured sparsity, we also select structured sparsity work as comparison baselines. Different from the previous experiments in this paper, we further take the input sparsity at each layer into account rather than only count the output sparsity. This is due to the fact that the baselines consider all zero operands. The results are listed in Table \ref{tab:comparison}, from which we can see that DSG is able to achieve a good balance between the operation amount and model accuracy.

\end{appendices}

\end{document}